\documentclass[letterpaper,twocolumn,10pt]{article}
\usepackage{usenix-2020-09}

\usepackage{tikz}
\usepackage{amsmath}

\usepackage{filecontents}
\usepackage{enumitem}
\usepackage{pbox}
\usepackage{graphicx}
\usepackage{subcaption}
\usepackage{appendix}
\usepackage{amsthm}
\usepackage{float}
\usepackage{amsfonts}
\usepackage{multirow}
\usepackage[normalem]{ulem}
\usepackage[linesnumbered,ruled]{algorithm2e}

\newtheorem{problem}{Problem}
\newtheorem{theorem}{Theorem}
\newtheorem{lemma}{Lemma}
\newtheorem{definition}{Definition}

\newcommand{\name}{\textsc{Zen}}

\newtheorem{innercustomthm}{Theorem}
\newenvironment{customthm}[1]
  {\renewcommand\theinnercustomthm{#1}\innercustomthm}
  {\endinnercustomthm}

\newtheorem{innercustomlemma}{Lemma}
\newenvironment{customlemma}[1]
  {\renewcommand\theinnercustomlemma{#1}\innercustomlemma}
  {\endinnercustomlemma}

\begin{document}
%-------------------------------------------------------------------------------

% \pagenumbering{gobble}

%don't want date printed
\date{}

% \title{\Large \bf {\name}: Near-Optimal Communications for Sparse and Distributed DNN Training}

% \title{\Large \bf {\name}: Near-Optimal Sparse Tensor Synchronization for Distributed DNN Training}

% \title{\Large On the Impact of Sparsity on Data Synchronization \\with Application to Distributed Training}

\title{\Large Empowering Distributed Training with Sparsity-driven Data Synchronization}

%for single author (just remove % characters)
% \author{
% {\rm Zhuang Wang*, ~Zhaozhuo Xu*, ~Anshumali Shrivastava,~and T. S. Eugene Ng} \\
% {\rm Rice Universtiy}
% \and
% {\rm Zhaozhuo Xu}\\
% Rice Universtiy
% copy the following lines to add more authors
% \and
% {\rm Name}\\
%Name Institution
% } % end author
% \author{
% {\rm Paper ID: 592, 12 pages body + references and separated appendix}
% } % end author

%for single author (just remove % characters)
\author{
{\rm Zhuang Wang\footnotemark[1]} \\
Rice University
\and
{\rm Zhaozhuo Xu\footnotemark[1]} \\
Stevens Institute of Technology
\and
{\rm Jingyi Xi\footnotemark[1]} \\
Zhejiang University
\and
{\rm Yuke Wang} \\
Rice University
\and
{\rm Anshumali Shrivastava} \\
Rice University
\and
{\rm T. S. Eugene Ng} \\
Rice University
} 

% \authornote{Authors contributed equally.}

\maketitle
\def\thefootnote{*}\footnotetext{These authors contributed equally.}\def\thefootnote{\arabic{footnote}}

%-------------------------------------------------------------------------------
\vspace{-0.1in}
\begin{abstract}

% \textcolor{red}{For the title of the paper, I'm thinking maybe we need to elevate section 2's contribution. Something like understanding the implications of sparse tensors on communication efficiency. Will keep thinking.}
%------------------------------------------------------------------------------
Distributed training is the de facto standard to scale up the training of deep learning models with multiple GPUs.
Its performance bottleneck lies in communications for gradient synchronization. 
Although high tensor sparsity is widely observed, the optimal communication scheme to fully leverage sparsity is still missing.
This paper aims to bridge this gap.
We first analyze the characteristics of sparse tensors in popular models to understand the fundamentals of sparsity. 
We then systematically explore the design space of communication schemes for sparse tensors and find the optimal ones.
These findings give a new understanding and inspire us to develop a holistic gradient synchronization system for sparse tensors called {\name}.
We demonstrate that {\name} can achieve up to $5.09\times$ speedup in communication time and up to $2.48\times$ speedup in training throughput compared to the state-of-the-art methods. \name\footnote{\url{https://anonymous.4open.science/r/zen-99D5}}~is also opensource anonymously.
\end{abstract}
%-------------------------------------------------------------------------------

%-------------------------------------------------------------------------------
\vspace{-0.15in}
\section{Introduction}
\vspace{-0.1in}

In recent years, deep learning (DL) models have achieved remarkable empirical performance in real-world applications, such as language processing~\cite{kiros2015skip, luong2015effective, vaswani2017attention, achiam2023gpt4, team2023gemini, dubey2024llama} and recommendation systems~\cite{guo2017deepfm, medini2020solar}.
With the ever-growing size of models and training datasets, distributed training has become the norm for model training with multiple GPUs~\cite{pytorch_ddp, horovod, pipedream, megatronlm, wang-gemini}.
% In DDT, the training set is partitioned among GPUs. 
% Next, starting from the same DNN initialization, each GPU performs mini-batch training on DNN using its own data. The DNN gradients generated by each GPU in every iteration are synchronized. Finally, the synchronized gradient is used to update the DNN in each GPU. 
%
The synchronization of gradient tensors from different GPUs is commonly required in distributed training.
% \sout{In data parallelism~\cite{horovod, byteps_osdi, pytorch_ddp}, because the training dataset is partitioned among GPUs, their  generated gradient tensors need to be synchronized to ensure model consistency.}
{In data parallelism~\cite{horovod, byteps_osdi, pytorch_ddp}, the training dataset is partitioned across multiple GPUs, so the corresponding gradient tensors must be synchronized to maintain model consistency.}
In tensor parallelism~\cite{shoeybi2019megatron}, individual layers of a DL model are sharded over multiple GPUs, which synchronize the gradient tensors during backward propagation for the gradient computation of subsequent layers.
Additionally, it is common practice to train large models~\cite{megatronlm, chowdhery2022palm, scao2022bloom} with a mix of data parallelism and other parallelism strategies, such as pipeline parallelism~\cite{pipedream, huang2019gpipe}, tensor parallelism~\cite{shoeybi2019megatron}, and ZeRO~\cite{rajbhandari2020zero, fsdp}.
These training workloads must synchronize gradient tensors across GPUs as well.

The major efficiency bottleneck of distributed training lies in communication for gradient synchronization~\cite{HiPress, wang2023cupcake, wang2023hi, zhang2022mics}.
Recent hardware developments have greatly improved the computation efficiency of model training. 
These advancements increase the frequency of gradient synchronization in distributed training and shift more burdens to the network systems.
However, the network upgrades have not kept up with computation improvements~\cite{Mellanox_update, OpenAI, nvidia_performance, zhang2022mics}, exacerbating the tension between computation and communication. 

Several communication schemes have been developed to alleviate the communication bottlenecks in distributed training by fully utilizing the network bandwidth.
For example, Ring-Allreduce~\cite{ring-allreduce2009} is provably bandwidth-optimal in homogeneous GPU clusters and it is widely used in collective communication libraries such as MPI~\cite{mpi_standard}, NCCL~\cite{NCCL}, and MSCCL~\cite{msccl}. 
BytePS~\cite{byteps_osdi} is a communication-optimal architecture for heterogeneous GPU/CPU clusters. 
However, both Ring-Allreduce and BytePS assume tensors to be synchronized are dense, ignoring their sparsity.
% the synchronization involves all gradients in a tensor, i.e., \textit{dense tensor}, regardless of the tensor sparsity.

Recent works have shown that DL models exhibit a high degree of tensor sparsity during gradient synchronization. 
The sparsity comes from either natural gradient computation~\cite{alizadeh2023llm, medini2020solar,clp+21,li2022embrace} or {gradient compression sparsification algorithms~\cite{aji2017sparse, wang2023cupcake, m2021efficient, li2022near}}.
On the one hand, because the model training may focus on updating a subset of parameters instead of all of them~\cite{frankle2018lottery,you2020drawing}, some of the gradient tensors in DL models are naturally sparse, with most of the gradients being zeros.
%
\iffalse
Table~\ref{tab:model_sparsity} lists the statistics of three \textcolor{red}{foundational}(\sout{widely deployed}) DL models for recommendation systems and language processing.
%
We can see that the embedding layers comprise a large portion of model parameters and show a significant sparsity level in their gradient tensors, with only around $2\%$ non-zero gradients.
\fi
% LSTM, guo2017deepfm, luong2015effective
{
% Modern DL models come with abundant natural tensor sparsity. 
For instance, the embedding table in the widely deployed Deep Learning Recommendation Models~\cite{wang2024oper} can reach more than 93\% sparsity; Graph adjacency matrix in Graph Neural Networks~\cite{wang2023tc} usually retains the sparsity more than 99\%; Word embeddings of those popular natural language processing models, such as NMT~\cite{luong2015effective} and LSTM~\cite{LSTM}, would achieve more than 97\% sparsity. 
% Attention Matrix in LLMs~\cite{zhang2023h2o} has more than 95\% sparsity.
}
On the other hand, to address the communication bottleneck, a plethora of sparsification algorithms~\cite{aji2017sparse, dgc, stich2018sparsified, wang2023cupcake, topk, gtopk, gu2024omniccl, li2022near} are proposed to reduce the synchronization traffic by selecting a subset of the original stochastic gradients.
They can save up to 99\% of the gradient exchange while maintaining model accuracy~\cite{sahu2021rethinking, HiPress, dgc}.

Leveraging the notable sparsity can significantly reduce the traffic volume for gradient synchronization and shorten the communication time in distributed training.
We can represent non-zero gradients with a sparse format and denote the tensor with a sparse format as a \textit{sparse tensor}.
Previous works, such as AGsparse~\cite{pytorch_ddp}, SparCML~\cite{sparcml}, and OmniReduce~\cite{omnireduce}, have acknowledged this potential.
They use various sparse formats and synchronization schemes.
However, these approaches do not fully consider the fundamental characteristics of sparsity in DL models and lack an understanding of the optimal scheme, resulting in suboptimal communication performance for gradient synchronization.
%The problem this paper addressed is
To this end, this paper addresses the research problem: \textit{What is the optimal communication scheme for gradient synchronization of sparse tensors in distributed training?}

\iffalse
To tackle this question, it is essential to revisit the fundamentals of sparsity in DL models.
This paper thoroughly analyzes sparse tensors, profiling their characteristics in popular models~\cite{guo2017deepfm, jozefowicz2016exploring, luong2015effective, dubey2024llama, opt175b, team2024gemma} across GPUs and iterations. 
Insights are sought into the relationship between sparse tensors and different training workloads.
Furthermore, the study delves into the changes observed in sparse tensors before and after aggregation with varying GPU configurations. 
It also examines the locations and distributions of non-zero gradients in tensors.
\fi

To tackle this question, it is essential to revisit the fundamentals of DL models' sparsity.
%The key is to
We first unveil the characteristics of sparse tensors in popular models~\cite{guo2017deepfm, jozefowicz2016exploring, luong2015effective, dubey2024llama, opt175b, team2024gemma} across GPUs, iterations, and diverse training workloads. 
We then explore locations and distributions of non-zero gradients in tensors and the changes observed in sparse tensors before and after aggregation under varying configurations.
In light of these characteristics of sparse tensors, we systematically explores the design space of synchronization schemes for sparse tensors.
Four elemental dimensions, including communication, aggregation, partition, and balance, are proposed to construct diverse synchronization schemes, revealing that existing schemes~\cite{pytorch_ddp, sparcml, omnireduce} can be described within this framework. 
This leads to the proof of the existence of communication-optimal schemes for synchronizing sparse tensors in distributed training.
\begin{figure}[t]
    \centering
	\includegraphics[width=\linewidth]{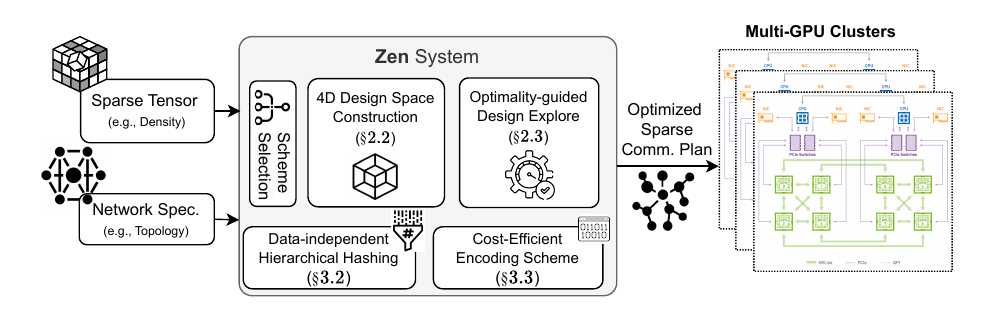}
	\vskip -0.1in
	\caption{\name~System Overview.}
	\label{fig:overview}
        \vspace{-0.2in}
\end{figure}

Building on these findings, we develop a sparse gradient synchronization system, called \name~(Figure~\ref{fig:overview}), achieving near-optimal communication time. {Specifically, \name~takes the tensor sparsity and network specification as the input and pinpoints the optimized sparse communication plan as the output.}
% {Our key design insight is to} frame the problem of realizing optimal schemes as a mathematical formulation and propose a data-independent solution to solve this problem, eliminating costly data-dependency overhead.
Our key design insight lies in framing the challenge of achieving optimal schemes as a mathematical problem. We propose a data-independent solution to address this challenge, eliminating costly data-dependency overhead.
This solution achieves high efficiency by introducing a novel hierarchical hashing algorithm that attains superior performance on GPUs while ensuring provably near-optimal results without any information loss.
{\name} also incorporates an efficient encoding scheme to minimize index representation overhead, irrespective of the random distribution of non-zero gradients after applying the hierarchical hashing algorithm.
We evaluate {\name} across various DL training workloads, where tensor sparsity arises from either natural computation or gradient compression.
{\name} achieves up to $5.09\times$ speedup in communication time and up to $2.48\times$ speedup in training throughput over existing methods for sparse tensor synchronization.

% The rest of this paper is organized as follows. 
% Section~\ref{sec:analysis} analyses the characteristics of sparse tensors, explores the design space of synchronization
% schemes for sparse tensors, and finds the optimal schemes.
% Section~\ref{sec:zen} introduces {\name}, the novel gradient synchronization system for sparse tensors.
% Section~\ref{sec:evaluation} provides experimental results on the performance of the proposed system.
% Section~\ref{sec:related_work} summarizes the state-of-the-art of related studies and Section~\ref{sec:conclusion} concludes this paper.

{The key contributions of this paper are as follows.
\begin{itemize}
    \vspace{-0.08in}
    \item We analyze the characteristics of sparse tensors in popular DL models to reveal and understand the fundamentals of gradient sparsity in distributed training. 
    \vspace{-0.08in}
    \item We systematically explore the design space of sparse communication schemes and pinpoint the optimal ones.
    \vspace{-0.08in}
    \item We develop {\name}, a gradient synchronization system for sparse tensors that achieves near-optimal communication time using a data-independent hierarchical hashing algorithm and an efficient encoding scheme.
    \vspace{-0.08in}
    \item Evaluation reveals that \name~achieves evident throughput improvement compared to the state-of-the-art methods across diverse training workloads.
\end{itemize}}
%-------------------------------------------------------------------------------

%-------------------------------------------------------------------------------
% \input{background}
%-------------------------------------------------------------------------------

%-------------------------------------------------------------------------------
\vspace{-0.15in}
\section{Analysis of Sparse Tensor Synchronization}
\label{sec:analysis}
% \vspace{-0.1in}
% \subsection{Gradient Sparsity in Deep Learning Training}
% \vspace{-0.05in}
\label{sec:motivate_sparsity}

% \begin{table*}[t!]
% \footnotesize
% \vspace{0.04in}
% \centering
% \begin{tabular}{ccccccc}
% \hline
% Model  & Task & Dataset & MLP Gradient Size & Embedding Gradient Size & Batch Size & Density  \\
% \hline \hline
% LSTM~\cite{LSTM}  & Language Modeling & One Billion Word & 20M & 406M & 128 &	 $1.13\%$ \\
% DeepFM~\cite{guo2017deepfm}  & Click-through Rate Prediction &   Criteo & 68M &	214M & 1024 &	 $2.80\%$ \\
% % NCF   & Collaborative Filtering & Yelp & 10M	& 132M 	& 16 & $0.06\%$		\\
% NMT~\cite{luong2015effective}  &Machine Translation & IWSLT 2014 De-En &	31M &	112M & 64 & $2.47\%$ 	\\
% % BERT~\cite{bert}  & Question Answering & SQuAD v1.1 & 86M	& 23M	& 4 &	$1.06\%$ \\
% \hline
% \end{tabular}
% \caption{Three Deep Learning models and their training statistics.}
% \label{tab:model_sparsity}
% \vspace{-0.2in}
% \end{table*}

\begin{table}[t]
\footnotesize
\caption{Three DL models with natural tensor sparsity.}
\vspace{-7pt}
% \vspace{0.05in}
\centering
\scalebox{0.98}{
\begin{tabular}{ccccccc}
\hline
\textbf{Model}  & \textbf{MLP Size} & \textbf{Embedding Size} & \textbf{Embedding Density}  \\
\hline \hline
LSTM~\cite{LSTM} & 20M & 406M &	 $1.13\%$ \\
DeepFM~\cite{guo2017deepfm} & 68M &	214M & 	 $2.80\%$ \\
% NCF   & Collaborative Filtering & Yelp & 10M	& 132M 	& 16 & $0.06\%$		\\
NMT~\cite{luong2015effective} &	31M &	112M & $2.47\%$ 	\\
% BERT~\cite{bert}  & Question Answering & SQuAD v1.1 & 86M	& 23M	& 4 &	$1.06\%$ \\
\hline
\end{tabular}}
\label{tab:model_sparsity}
\vspace{-5pt}
\end{table}

\begin{table}[t!]
\footnotesize
\centering
\caption{Configurations of three LLMs. AH: attention heads.}
\vspace{-7pt}
\scalebox{0.98}{
\begin{tabular}{ccccc}
\hline \hline
\textbf{Model} & \textbf{Hidden size} & \textbf{Intermediate} & \textbf{\#Layers} & \textbf{\#AH} \\
\hline
Llama3.2-3B\cite{dubey2024llama} & 3,072 & 8,192 & 28  & 24 \\
OPT2.7B\cite{opt175b} & 2,560 & 10,240 & 32 & 32 \\
Gemma2-2B\cite{team2024gemma} & 2,304 & 9,216 & 26 & 8 \\
\hline
\end{tabular}}
\label{tab:llm_configs}
\vspace{-10pt}
\end{table}

\vspace{-0.05in}
\subsection{Characteristics of Sparse Tensors}
\label{sec:char_sparse}
\vspace{-0.03in}
This section analyzes the characteristics of sparse tensors from both natural gradient computation and sparsification algorithms.
For natural tensor sparsity, we will study the gradient tensors from embedding layers of the three DL models listed in Table~\ref{tab:model_sparsity}.
For tensor sparsity from gradient compression, we will study three large language models (LLMs) listed in Table~\ref{tab:llm_configs} and apply DGC~\cite{dgc}, one of the most popular sparsification algorithms, to select the top 5\% gradients from their gradient tensors~\cite{gc_llm2024, gc_llm_mlsys}.
The detailed workloads for the models can be found in \S\ref{exp_setup}.

% The original gradient tensor is in a dense format, in which the gradients of all the parameters in a Deep Learning layer are stored.
\vspace{-0.03in}
\begin{definition}[Dense tensor]
We define the original gradient tensor in a DL layer as a dense tensor.
\end{definition}

\vspace{-0.03in}
We define \textit{density} of a gradient tensor as the percentage of its non-zero gradient values.
We can represent a gradient tensor in a sparse format when many parameters have zero gradients. 
A typical sparse format is coordinate lists (COO) that store a list of non-zero gradients and a list of the corresponding indices~\cite{omnireduce, xu2021grace}.

\vspace{-0.03in}
\begin{definition}[Sparse tensor]
We define a gradient tensor in a sparse format as a sparse tensor.
\end{definition}

\vspace{-0.03in}
We denote the size of a dense tensor $G$ is $M$, its density is $d_G$, and the training involves $n$ nodes.
For simplicity, we assume each node has only one GPU in this section.

% \begin{figure}[t!]
%     \centering
% 	\begin{subfigure}[t]{0.26\linewidth}
% 	\includegraphics[width=\linewidth]{figures/fully_overlap.pdf}
% 	\vspace{-0.2in}
% 	\caption{Full overlap}
% 	\label{fig:fully_overlap}
%     \end{subfigure} \hspace{0.04\linewidth}
%     \begin{subfigure}[t]{0.26\linewidth}
% 	\includegraphics[width=\linewidth]{figures/partial_overlap.pdf}
% 	\vspace{-0.2in}
% 	\caption{Partial overlap}
% 	\label{fig:partial_overlap}
%     \end{subfigure} \hspace{0.04\linewidth}
%     \begin{subfigure}[t]{0.26\linewidth}
% 	\includegraphics[width=\linewidth]{figures/no_overlap.pdf}
% 	\vspace{-0.2in}
% 	\caption{No overlap}
% 	\label{fig:no_overlap}
%     \end{subfigure} 
%     \caption{Examples of the overlaps between two sparse tensors. The numbers are the indices of the gradients. How much they can overlap is unknown a priori.}
%     \label{fig:overlaps}
% \end{figure}

\vspace{3pt}
\noindent 
\textbf{C1: The overlap of sparse tensors varies.}
Similar to dense tensors, sparse tensors are aggregated during synchronization.
When aggregating dense tensors, the indices of gradients from different GPUs are identical.
However, due to different batches as the input for training on different GPUs, the indices of non-zero gradients in a sparse tensor are unknown \textit{a priori}.
They can have overlaps, but how much they overlap depends on many factors, such as the DL model, the training dataset, and the batches.
% and Figure~\ref{fig:overlaps} displays three cases for the overlaps between two sparse tensors.
% When two sparse tensors are fully overlapped, they have the same set of indices for non-zero gradients.
% When two sparse tensors have no overlaps, their sets of indices are disjoint.
% A more common case is that the sets of indices are partially overlapped.
We define the overlap ratio~\cite{vijaymeena2016survey} to quantify this overlap between two sparse tensors.

\vspace{-0.06in}
\begin{definition}[The overlap ratio] \label{def:overlap_ratio}
Given two sparse tensors and their sets of indices for non-zero gradients are $I_1$ and $I_2$, respectively, their overlap ratio is defined as $\frac{|I_1 \cap I_2|}{\min \{|I_1|, |I_2|\}}$, where $|\cdot|$ is the cardinality of a set.
\end{definition}
\vspace{-0.06in}

\begin{figure}[t] \small
    % \vspace{0.05in}
    \centering
	\begin{subfigure}[t]{0.5\linewidth}
	\includegraphics[width=0.9\linewidth]{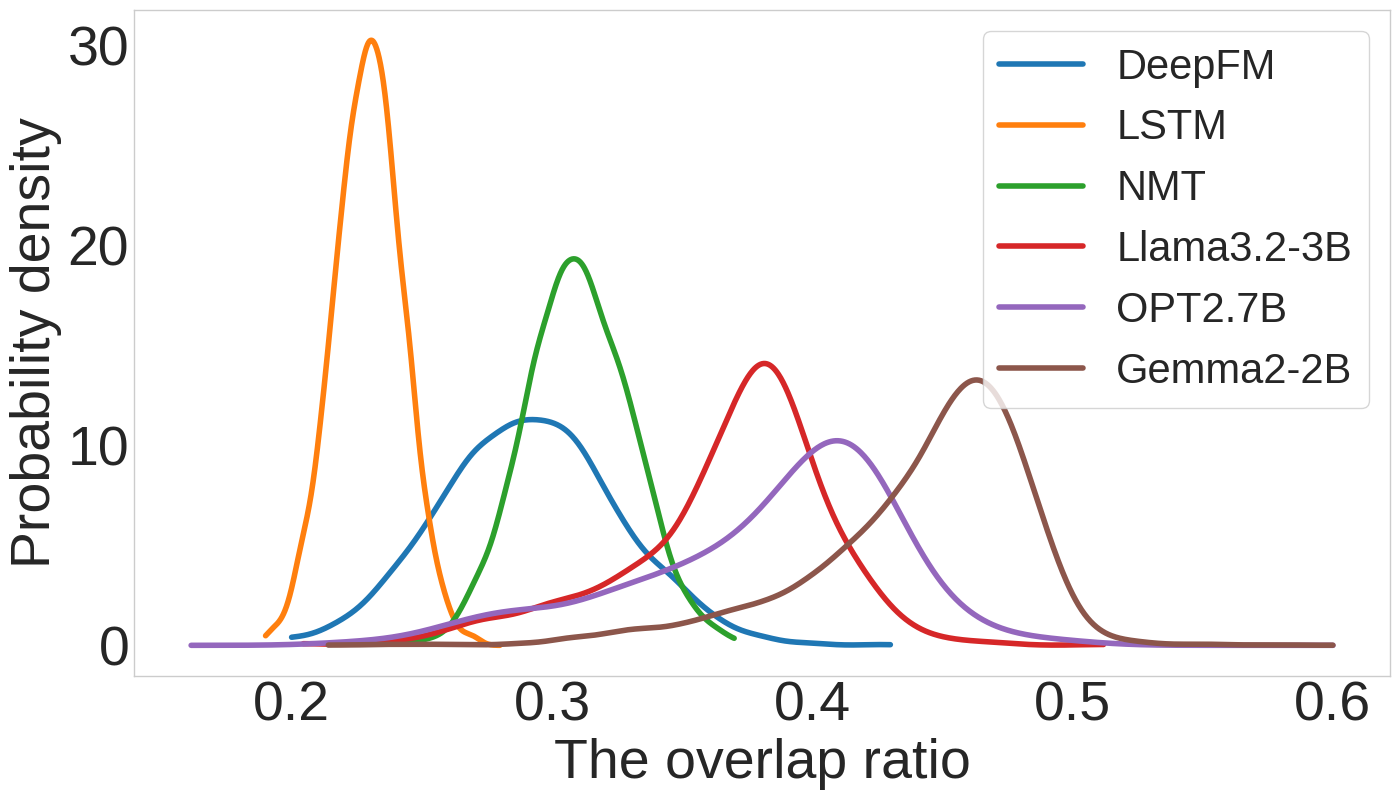}
	\vspace{-0.1in}
	\caption{}
	\label{fig:overlap_ratio}
    \end{subfigure}  
    \begin{subfigure}[t]{0.46\linewidth}
	\includegraphics[width=0.9\linewidth]{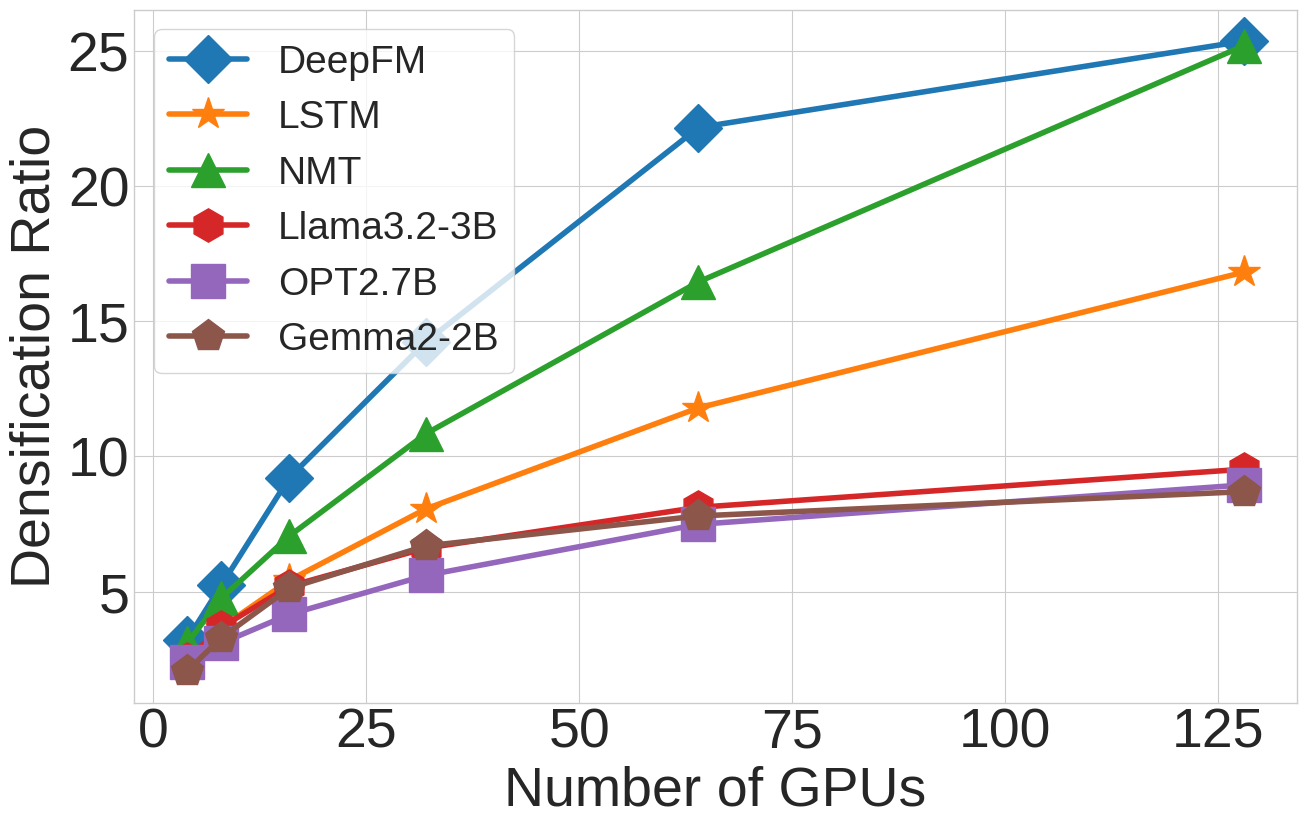}
	\vspace{-0.1in}
	\caption{}
	\label{fig:denser_ratio}
    \end{subfigure} 
    \vskip -0.05in
    \caption{The characteristics of sparse tensors in DL models. (a) shows that the overlap ratio of sparse tensors varies; (b) shows that tensors have higher density after aggregation.}
    \label{fig:overlap}
    % \vspace{-0.1in}
\end{figure}

Figure~\ref{fig:overlap_ratio} shows the probability density function (PDF) of the overlap ratios for the six studied DL models.
We can see that the overlap ratio in a model is approximately normally distributed and it is in a wide range.
% For example, the overlap ratio in DeepFM is in the range of $[0.21, 0.44]$.
In addition, different models have different distributions of overlap ratios. 
% For example, the range of the overlap ratio in BERT is $[0.11, 0.29]$, while in NMT is $[0.24, 0.37]$.

\begin{figure}[t]
    \centering
	\begin{subfigure}[t]{0.48\linewidth}
	\includegraphics[width=0.9\linewidth]{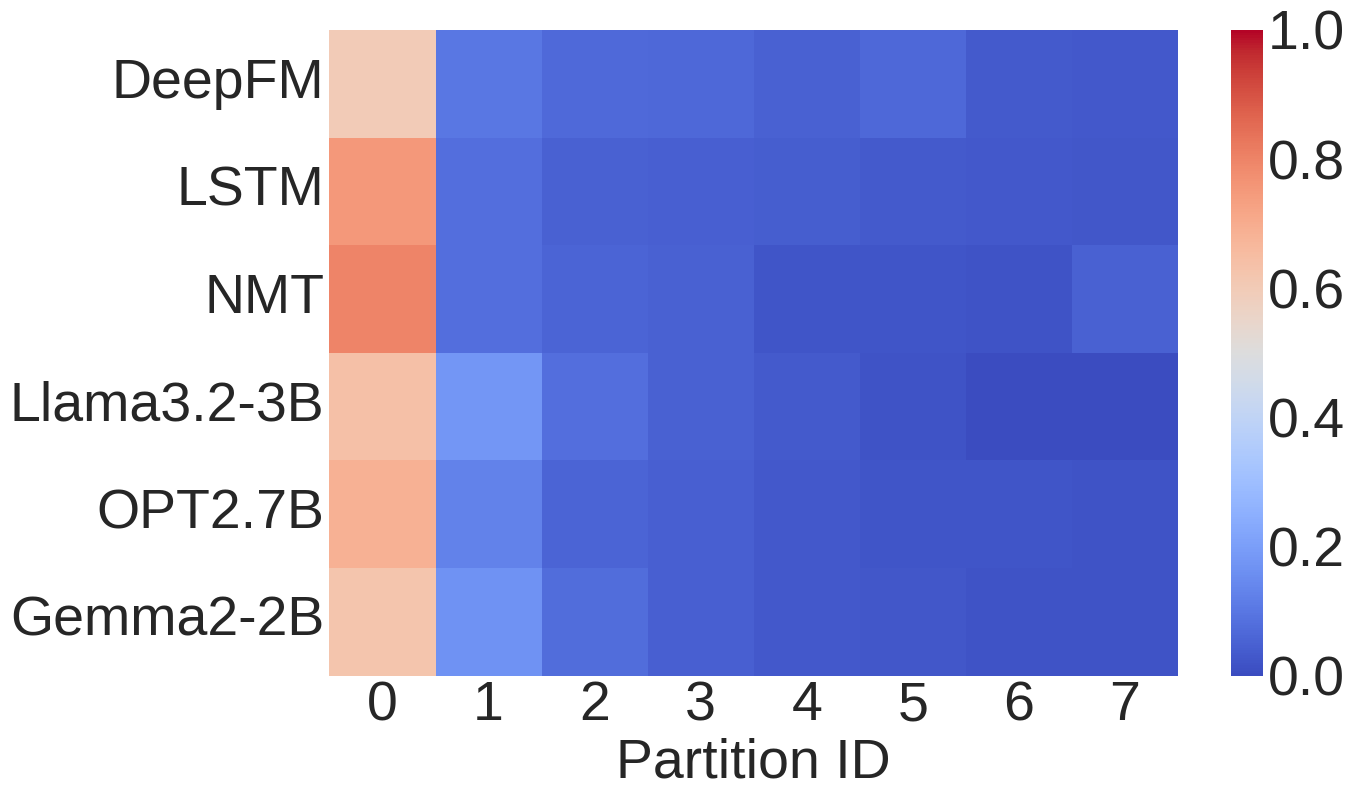}
	\vspace{-0.1in}
	\caption{}
	\label{fig:heatmap}
    \end{subfigure}  
    \begin{subfigure}[t]{0.47\linewidth}
	\includegraphics[width=0.9\linewidth]{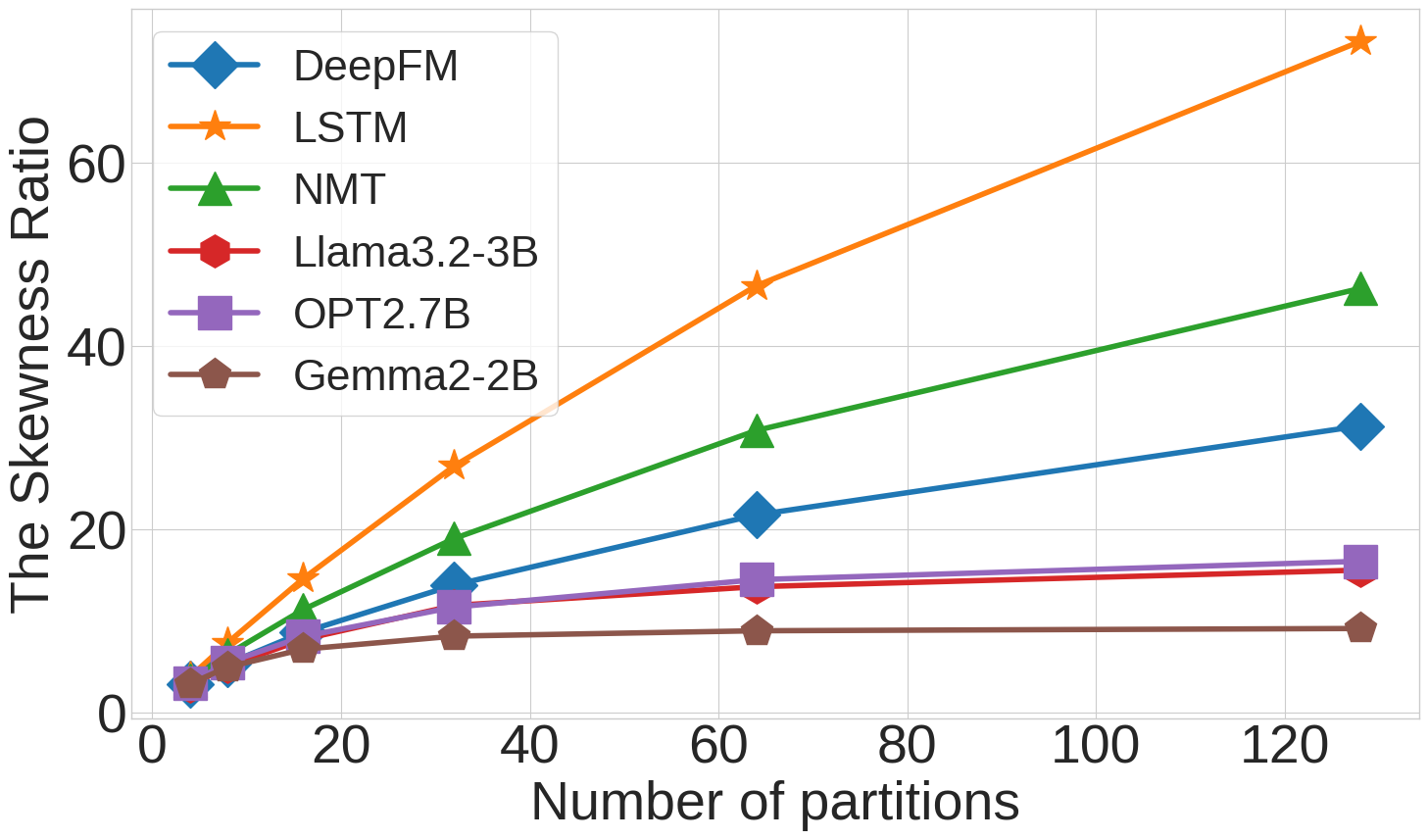}
	\vspace{-0.1in}
	\caption{}
	\label{fig:skew_ratio}
    \end{subfigure} 
    \vskip -0.05in
    \caption{The distribution of non-zero gradients is skewed. (a) The heatmap of non-zero gradients distribution; (b) The skewness ratio. }
    \label{fig:skewness}
    \vskip -0.1in
\end{figure}

\vspace{3pt}
\noindent
\textbf{C2: The tensor size after aggregation varies.}
When aggregating dense tensors, the tensor sizes before and after aggregation remain the same.
However, when aggregating sparse tensors, the unknown overlaps of sparse tensors lead to varying tensor sizes after aggregation.
% As shown in Figure~\ref{fig:overlaps}, although the two sparse tensors have the same number of non-zero gradients in the three examples, the aggregated tensor sizes are different.
Because the aggregation involves sparse tensors from multiple GPUs, we denote $d_G^n$ as the density after the aggregation of tensors from $n$ GPUs.
We observe that sparse tensors get denser after aggregation. 
Here we define the densification ratio to quantify this characteristic.

\begin{definition}[The densification ratio] \label{def:denser_ratio}
Given a dense tensor $G$, its densification ratio is define as $\gamma_G^n = \frac{d_G^n}{d_G}$.
\end{definition}

Figure~\ref{fig:denser_ratio} presents the average densification ratio $\gamma_G^n$ to the number of GPUs for the DL models studied in this paper.
The densification ratio increases with the number of GPUs, demonstrating that tensors have higher density after aggregation.
We can also see that the densification ratio is smaller than the number of GPUs, i.e., $\gamma_G^n < n$.
It suggests that the indices of non-zero gradients in sparse tensors from different GPUs are partially overlapped.

\begin{figure}[t]
    % \vspace{0.05in}
    \centering
	\begin{subfigure}[t]{0.22\linewidth}
	\includegraphics[width=\linewidth]{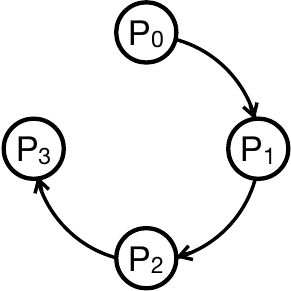}
	\vspace{-0.2in}
	\caption{Ring.}
	\label{fig:ring}
    \end{subfigure}  \hspace{0.02\linewidth}
    \begin{subfigure}[t]{0.32\linewidth}
	\includegraphics[width=\linewidth]{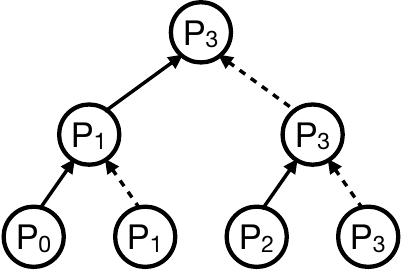}
	\vspace{-0.2in}
	\caption{Hierarchy.}
	\label{fig:hierarchy}
    \end{subfigure} \hspace{0.02\linewidth}
    \begin{subfigure}[t]{0.27\linewidth}
	\includegraphics[width=0.8\linewidth]{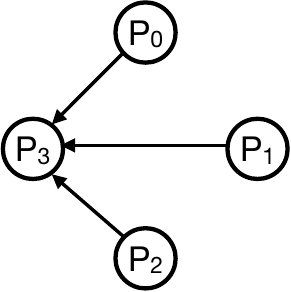}
	\vspace{-0.04in}
	\caption{Point-to-point.}
	\label{fig:broadcast}
    \end{subfigure} 
    % \vskip -0.1in
    \caption{An illustration of three communication patterns with four GPUs. GPU $P_3$ aggregates the data from all GPUs.}
    \label{fig:comm_patterns}
    % \vskip -0.1in
\end{figure}

\begin{figure}[t] \small
    \centering
    \begin{subfigure}[t]{0.42\linewidth}
	\includegraphics[width=0.9\linewidth]{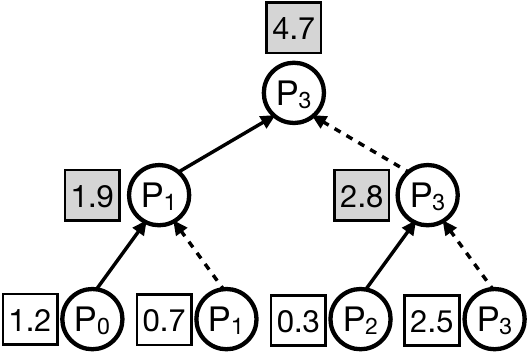}
	\vspace{-0.05in}
	\caption{Incremental aggregation.}
	\label{fig:agg_incremental}
    \end{subfigure} \hspace{0.05\linewidth}
    \begin{subfigure}[t]{0.42\linewidth}
	\includegraphics[width=0.9\linewidth]{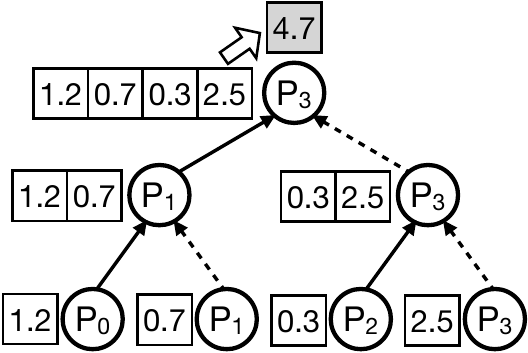}
	\vspace{-0.05in}
	\caption{One-shot aggregation.}
	\label{fig:agg_one_shot}
    \end{subfigure} 
    \vskip -0.08in
    \caption{An Illustration of two aggregation patterns with Hierarchy. The gradients on each GPU are from the same parameter and 4.7 is the final aggregated result.}
    \label{fig:agg_patterns}
    % \vskip -0.1in
\end{figure}

\begin{figure*}[t]
% \vspace{0.02in}
    \centering
	\begin{subfigure}[t]{0.32\linewidth}
	\includegraphics[width=0.9\linewidth]{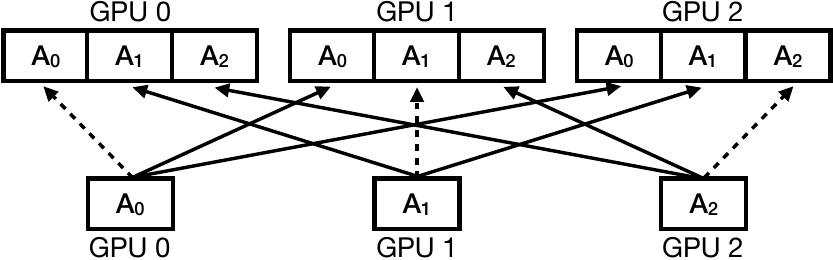}
	\vspace{-0.02in}
	\caption{Centralization}
	\label{fig:centralization}
    \end{subfigure}  
    \hspace{0.1\linewidth}
    \begin{subfigure}[t]{0.45\linewidth}	\includegraphics[width=0.9\linewidth]{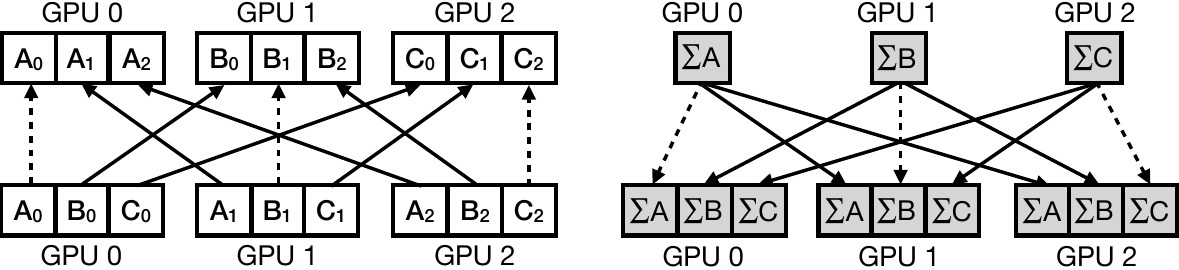}
	\vspace{-0.02in}
	\caption{Parallelism}
	\label{fig:parallelism}
    \end{subfigure} 
    \vskip -0.1in
    \caption{An illustration of the two partition patterns with Point-to-point. In (a), each tensor is communicated as a whole and each GPU receives all the tensors. In (b), each tensor is split into three partitions; the same partition from different GPUs is sent to the same place, and the aggregated results are then sent back to all GPUs.}
    \label{fig:partition}
\end{figure*}
\begin{figure*}[t]
    \centering
	\begin{subfigure}[t]{0.4\linewidth}
	\includegraphics[width=0.9\linewidth]{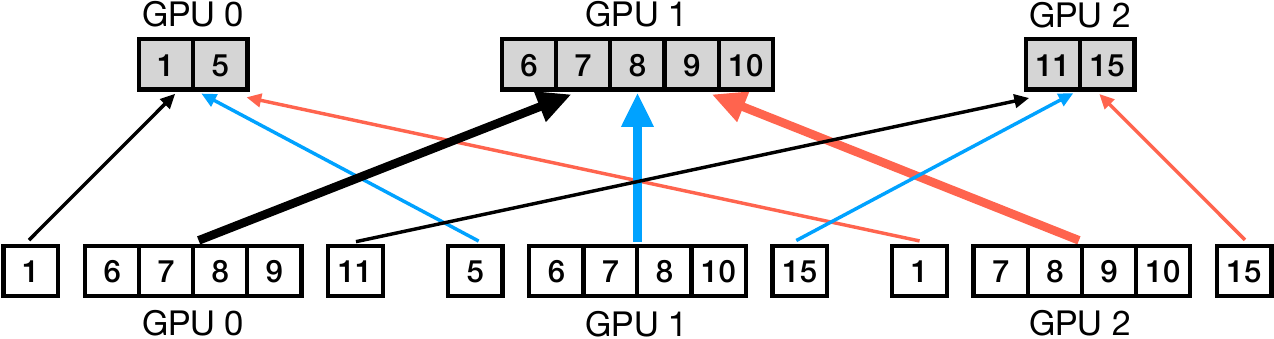}
	\vspace{-0.05in}
	\caption{Imbalanced communication}
	\label{fig:imbalanced_comm}
    \end{subfigure}  
    \hspace{0.06\linewidth}
    \begin{subfigure}[t]{0.4\linewidth}
	\includegraphics[width=0.9\linewidth]{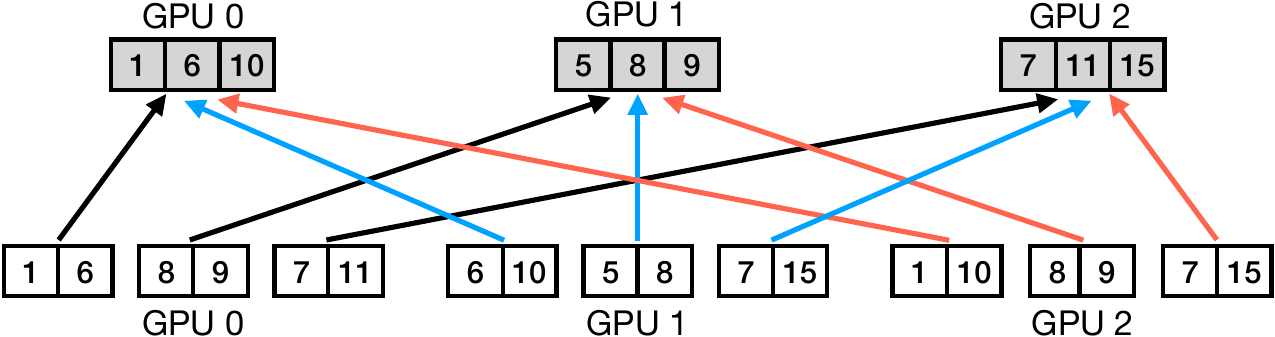}
	\vspace{-0.05in}
	\caption{Balanced communication}
	\label{fig:balanced_comm}
    \end{subfigure} 
    \vskip -0.1in
    \caption{An illustration of balance patterns with Point-to-point and Parallelism. Each GPU has six non-zero gradients and the numbers are their indices.  In (a), four gradients from each GPU are sent to GPU 1. In (b), each GPU sends two gradients to other GPUs, and communications are well-balanced. However, it is non-trivial to achieve such balanced communications.}
    \label{fig:balance}
    \vskip -0.1in
\end{figure*}

\vspace{3pt}
\noindent
\textbf{C3: The distribution of non-zero gradients is skewed.}
% The distribution of the indices for gradients in dense tensors is uniform because every gradient is communicated for synchronization, no matter its value is zero or not.
% However, the distribution of indices for non-zero gradients is skewed.
When evenly splitting a dense tensor into multiple partitions, we observe that most of the non-zero gradients are in one of them.
For example, with eight partitions, over 60\% of the non-zero gradients are in the first partition in the six DL models.
Figure~\ref{fig:heatmap} shows the percentage heatmap of the non-zero gradients in each partition.
Here we define the skewness ratio to quantify the skewed distribution of non-zero gradients.

\begin{definition}[The skewness ratio]\label{def:skew_ratio}
Given a dense tensor $G$ and we evenly split $G$ into $n$ disjoint partitions, denoted as $\{G_1, \cdots, G_n\}$, then the skew ratio of $G$ with $n$ partitions is defined as $\frac{\max_{i \in [n]}\{d_{G_i}\}}{d_G}$.
\end{definition}

Figure~\ref{fig:skew_ratio} presents the skewness ratios of gradient tensors in the DL models studied in this paper.
They are significant in all six models. 
For example, when we evenly split the gradient tensor from the embedding table in LSTM into 128 partitions, the skewness ratio is over 70.
It indicates that more than half of the non-zero gradients are in the same partition.
% Another observation is that the skewness ratio consistently increases with the number of partitions.
% It suggests that the distribution of non-zero gradients gets "skewer" with more partitions. 

% \subsection{Performance of Synchronization Schemes}
% \label{sec:comm_problems}

% % \textcolor{red}{We could point to characteristics C1-C3 during the following to link the conclusions to the characteristics to establish a stronger connection}

% Synchronization schemes for dense tensors have been extensively studied~\cite{MPIoptimization, horovod, byteps_osdi, PSosdi2014}.
% In this section, we will first explore the design space to construct synchronization schemes for sparse tensors 
% % for the first time 
% (\S\ref{sec:ele_patterns}).
% We then find the optimal scheme according to the three characteristics of sparse tensors (\S\ref{sec:performance_analysis}).
% We also conduct a numerical comparison among different schemes to demonstrate the superior performance of the optimal scheme (\S\ref{sec:numerical_comp}).

% \subsubsection{Elemental Dimensions for Synchronization}
% \label{sec:ele_patterns}

\vspace{-0.05in}
\subsection{Elemental Dimensions for Synchronization}
% \subsection{\textcolor{red}{Key} Dimensions for Synchronization}
\vspace{-0.03in}
\label{sec:ele_patterns}
Synchronization for dense tensors has been extensively studied~\cite{MPIoptimization, horovod, byteps_osdi, PSosdi2014}.
This section will explore the design space to construct synchronization schemes for sparse tensors.
Given a tensor, the outcome of its synchronization is that gradients with the same indices are aggregated and all GPUs have identical aggregated results.
We will discuss four dimensions that construct a synchronization scheme for sparse tensors.

\noindent
\textbf{Communication dimension.} 
% \paragraph{Communication dimension.}
There are typically three communication patterns for synchronization: 1) \texttt{Ring}, 2) \texttt{Hierarchy}, and 3) \texttt{Point-to-point}.
They are illustrated in Figure~\ref{fig:comm_patterns} with an example in which there are four GPUs and GPU $P_3$ aggregates the data from all GPUs.
In Ring, all GPUs form a ring structure.
$P_0$ first sends its data to $P_1$, which then passes the data along with its own data to $P_2$ and so on until $P_3$ receives all the data.
In Hierarchy, all GPUs form a hierarchical structure and $P_3$ is the root.
There are two stages in Figure ~\ref{fig:hierarchy}.
In the first stage, $P_0$ sends its data to $P_1$ and $P_2$ sends its data to $P_3$. 
In the second stage, $P_1$ sends the data from both its own and $P_0$ to $P_3$.
% Note that this is just one example of the possible hierarchical structures. 
In Point-to-point communication, the other three GPUs directly send data to $P_3$.

\vspace{3pt}
\noindent
\textbf{Aggregation dimension.}
A communication pattern can have multiple communication stages and there are two options for aggregation: 1) \texttt{Incremental aggregation}, i..e, aggregate tensors at each stage; and 2) \texttt{One-shot aggregation}, i.e., only aggregate tensors from all GPUs after the last stage.
These two options lead to different amounts of traffic volume in different communication stages because of C1 and C2 discussed in Section~\ref{sec:char_sparse}.
In the example illustrated in Figure~\ref{fig:comm_patterns}, Ring has three stages and Hierarchy has two stages.
% Although each GPU has one tensor for synchronization, it can host multiple tensors at each stage. 
Figure~\ref{fig:agg_patterns} displays an example with Hierarchy as the communication pattern.
When $P_1$ receives a tensor from $P_0$, it has two tensors.
$P_1$ can either aggregate the two tensors and send the aggregated result to $P_3$, as shown in Figure~\ref{fig:agg_incremental};
or it can just send the concatenated tensor to $P_3$, as shown in Figure~\ref{fig:agg_one_shot}.

\vspace{3pt}
\noindent
\textbf{Partition dimension.}
There are two partition patterns to ensure that all GPUs have the same aggregated results after synchronization: 1) \texttt{Centralization}, in which each tensor is communicated and aggregated as a whole; and 2) \texttt{Parallelism}, in which each tensor is decomposed into multiple partitions and each partition is communicated and aggregated separately.
Figure~\ref{fig:partition} compares the two partition patterns with Point-to-point as the communication pattern.
With Centralization, as shown in Figure~\ref{fig:centralization}, each GPU sends its tensor as a whole to other GPUs.
With Parallelism, as shown in Figure~\ref{fig:parallelism}, each GPU first decomposes its tensor into three partitions and it requires two steps for synchronization.
The first step aggregates the same partition from different GPUs in different places and the second step ensures that all GPUs have the aggregated results of all partitions.

\vspace{3pt}
\noindent
\textbf{Balance dimension.}
With Parallelism, the number of non-zero gradients in each partition can vary. 
Therefore, there are two patterns in terms of the traffic volume received at each GPU: 
1) \texttt{Balanced communication}, in which GPUs receive the same amount of data; and 2) \texttt{Imbalanced communication}, in which the traffic volumes received at different GPUs are greatly different.
Figure~\ref{fig:balance} compares the two balance patterns among three GPUs with Point-to-point.
There are 15 gradients in the tensor and six of them are non-zero. 
As shown in Figure~\ref{fig:imbalanced_comm}, four non-zero gradients are in the middle partition and they are sent to GPU 1.
The traffic volume received at GPU 1 is $4\times$ that received at GPU 0 and GPU 2.
In Figure~\ref{fig:balanced_comm}, each GPU sends two non-zero gradients to other GPUs and the volume among them is well-balanced.
% In Balanced communication with Parallelism, the density of each partition before aggregation is $d_{G}/n$ and we denote $\gamma_{G/n}^n$ as its densification ratio.
Naively partitioning a tensor can cause imbalanced communications due to C3 discussed in \S\ref{sec:char_sparse}.

The four dimensions describe the design space of synchronization schemes for sparse tensors. 
Table~\ref{table:comparison} classifies existing schemes~\cite{pytorch_ddp, sparcml, omnireduce} based on their dimensions.

\begin{table*}[t] \small
% \vspace{0.03in}
\centering
\footnotesize
\caption{Comparison of different synchronization schemes for sparse tensors based on their dimensions. }
\vspace{-5pt}
\scalebox{0.9}{\begin{tabular}{llllll}
\hline 
\textbf{Schemes}          & \textbf{Communication}  & \textbf{Aggregation}  & \textbf{Partition} & \textbf{Balance} & \textbf{Note} \\ \hline  \hline                                      
AGsparse~\cite{pytorch_ddp}         & Ring, Hierarchy, Point-to-point                   & One-shot           & Centralization    & N/A            & \multirow{2}{*}{\shortstack{Cannot fully use overlaps \\ to reduce traffic.}} \\
SparCML~\cite{sparcml}          & Hierarchy             & Incremental          & Centralization     & N/A            &                                                                     \\ \hline      
% Sparse PS~\cite{PSosdi2014}        & Point-to-point        & One-shot             & Parallelism       & Imbalanced           & \multirow{2}{*}{\shortstack{Suffer from imbalanced \\ communications.}}          \\
OmniReduce~\cite{omnireduce}      & Point-to-point        & One-shot             & Parallelism       & Imbalanced   &   Imbalanced communications                                                                  \\ \hline                               
Balanced Parallelism              & Point-to-point        & Incremental             & Parallelism       & Balanced        &     The optimal scheme we find. \\    
\hline  
\end{tabular}}
\vspace{-5pt}
\label{table:comparison}
\end{table*}

% \vspace{-0.05in}
\subsection{Optimal Synchronization Schemes}
\label{sec:performance_analysis}
\vspace{-0.03in}
%We will next analyze the optimal schemes within the design space described by the four dimensions to synchronize sparse tensors. We define a scheme called \textbf{Balanced Parallelism}.
Next, we analyze optimal synchronization schemes for sparse tensors based on the four design dimensions. 
For convenience, we introduce two special schemes: \textit{\textbf{Balanced Parallelism}} and \textit{\textbf{Hierarchical Centralization}}.

\begin{definition}[Balanced Parallelism]
It refers to the synchronization scheme characterized by [Point-to-point, Incremental aggregation, Parallelism, and Balanced communication].
\end{definition}

\begin{definition}[Hierarchical Centralization]
It refers to the synchronization scheme characterized by [Hierarchy, Incremental aggregation, and Centralization].
\end{definition}
\begin{theorem}[Optimal schemes]
\label{thm:optimal_scheme}
%When choosing a synchronization scheme to minimize communication time for sparse tensors, the optimal scheme is either Balanced Parallelism or the scheme with [Hierarchy, Incremental aggregation, and Centralization].
To minimize communication time for sparse tensors, the optimal synchronization scheme is either {Balanced Parallelism} or {Hierarchical Centralization}.
\end{theorem}

\begin{proof}
%We prove Theorem~\ref{thm:optimal_scheme} with two lemmas. 
Theorem~\ref{thm:optimal_scheme} is proven using two lemmas: \ref{lemma:parallelism} and \ref{lemma:centralization}. Here, we present the main intuitions, with the detailed proof available in Appendix~\ref{app:proof_th1}.
%Due to space limitations, we mainly introduce the intuitions here and the detailed proof can be found in Appendix~\ref{app:proof_th1}.

\begin{lemma}
\label{lemma:parallelism}
When the partition pattern is fixed to Parallelism, the optimal scheme is  {Balanced Parallelism}.
\end{lemma}

%the optimal scheme adopts  \textbf{Balanced Parallelism}

There are three intuitions for Lemma~\ref{lemma:parallelism}: 
%1) schemes with Balanced communication outperform schemes with Imbalanced communication;
1) Balanced communication outperforms imbalanced communication;
%2) Point-to-point communication outperforms both Ring and Hierarchy because it minimizes the amount of traffic volume for the unique gradients when the partition dimension is Parallelism;
2) Point-to-point communication outperforms both Ring and Hierarchy by minimizing the traffic volume of unique gradients in the Parallelism partition pattern.
and 3) Incremental aggregation outperforms One-shot aggregation as it reduces the traffic volume by aggregating the overlaps of sparse tensors. 

\begin{lemma} 
\label{lemma:centralization}
When the partition pattern is fixed to Centralization, the optimal scheme is Hierarchical Centralization.
\end{lemma}

There are two intuitions for Lemma~\ref{lemma:centralization}: 
%1) when the partition pattern is fixed to Centralization, there are only six candidates in the search space because there is no Balance dimension; 
1) when the partition pattern is fixed to Centralization, the search space is reduced to six candidates because the Balance dimension is not applicable.
%2) the scheme with [Hierarchy, Incremental aggregation, and Centralization] can minimize the amount of traffic volume for the overlapped gradients compared to other schemes with Centralization.
2) Among these candidates, Hierarchical Centralization minimizes the traffic volume for overlapped gradients.
We use an extreme case to demonstrate the second intuition.
Suppose a non-zero gradient with index $idx$ appears in all sparse tensors. 
In Point-to-point or One-shot aggregation, each GPU has to receive this gradient for $n-1$ times. 
In Ring, the gradient from each GPU is aggregated at every stage and forwarded to the next GPU, 
causing each GPU to receive the gradient 
$n-1$ times.
However, with Hierarchy and Incremental aggregation, each GPU only receives the gradient $\log n$ times.

% \vspace{-0.07in}
% \begin{lemma}
% \label{lemma:sparml_comm_time}
% The theoretical communication time of the scheme that adopts Hierarchy, Incremental aggregation, and Centralization is $\sum_{i=1}^{\log n} \gamma_G^{2^{i-1}} 2Md_G/b$.
% \end{lemma}
% \vspace{-0.07in}

% \Zhuang{
% This scheme has $\log n$ communication stages and the traffic volume each GPU receives in the $i_{th}$ stage is 
% $2 \gamma_G^{2^{i-1}}d_G M$.
% The total traffic volume each GPU receives for synchronization is $2 \sum_{i=1}^{\log n} \gamma_G^{2^{i-1}}d_G M$.
% Therefore, the theoretical communication time is $\sum_{i=1}^{\log n} \gamma_G^{2^{i-1}} 2Md_G/b$.
% }

Lemma~\ref{lemma:parallelism} and Lemma~\ref{lemma:centralization} imply Theorem~\ref{thm:optimal_scheme}.
\end{proof}
% \vspace{-0.1in}

\begin{table}[t] \small
\footnotesize
% \vspace{0.04in}
\centering
\caption{The symbols used in this paper.}
\vspace{-7pt}
\begin{tabular}{cl}
\hline
\textbf{Symbols}     & \textbf{Description}                                      \\ \hline
$n$         & The number of GPUs                               \\
$b$         & The network bandwidth                              \\
$G$         & The dense tensor                             \\
$M$         & The size of the dense tensor                 \\
$d_G$       & The density of $G$                             \\
$\gamma_G^k$ & The densification ratio of $G$ with $k$ GPUs \\ \hline
\end{tabular}
\vspace{-0.1in}
\label{table:symbols}
% \vspace{-0.2in}
\end{table}

\vspace{-0.2in}
\paragraph{Communication time analysis.}
Next, we will analyze the theoretical communication time of the two schemes mentioned in Theorem~\ref{thm:optimal_scheme}.
The symbols used in the analysis are listed in Table~\ref{table:symbols}.
We assume each node is equipped with a single GPU, and each pairs of nodes have bidirectional, direct connections~\cite{sparcml}. The theoretical communication time is defined as the transfer time of messages, $L/b$ , where $L$ is the message size and $b$ is the network bandwidth~\footnote{We ignore the latency of message transmission, as it is negligible compared to transfer time when using Hierarchy or Point-to-point.}. For this analysis, we adopt the COO sparse format. For simplicity, we assume that the tensor on each GPU has identical $d_G$ and the average densification ratio for all GPUs is $\gamma_G^n$. Additionally, the number of GPUs, $n$, is a power of 2.

$\bullet$~\textbf{Balanced Parallelism.}
Communication in this scheme involves two steps.
The traffic volume each GPU receives is $\frac{2(n-1)}{n} d_G M$  in the first step and $\frac{2(n-1)}{n} \gamma_G^n d_G M$ in the second step. 
The total traffic volume each GPU receives is $\frac{2(n-1)}{n} (\gamma_G^n +1) d_G M$. resulting in a communication time of:
\vspace{-0.08in}
\begin{equation}
    T_{bp} = \frac{n-1}{n} (\gamma_{G}^n+1) 2Md_G/b. 
\label{eq:bp}
\end{equation}

$\bullet$~\textbf{Hierarchical Centralization.}
communication in this scheme is performed in $logn$ stages. In the $i_{th}$ stage, the traffic volume each GPU receives is 
$2 \gamma_G^{2^{i-1}}d_G M$.
The total traffic volume each GPU receives for is $2 \sum_{i=1}^{\log n} \gamma_G^{2^{i-1}}d_G M$.
The corresponding communication time is:
\begin{equation}
     T_{sm} = \sum_{i=1}^{\log n} \gamma_G^{2^{i-1}} 2Md_G/b.
\label{eq:sm}
\end{equation}

% \vspace{-0.15in}
\paragraph{Balanced Parallelism as the practical optimal scheme.}
According to Theorem~\ref{thm:optimal_scheme}, the optimal scheme is determined by comparing Equations~(\ref{eq:bp}) and~(\ref{eq:sm}).
Based on the characteristics of sparse tensors in DL models, we observe that Balanced Parallelism is often better than another scheme because $\sum_{i=1}^{\log n} \gamma_G^{2^{i-1}} > \frac{n-1}{n} (\gamma_{G}^n+1)$ in practical distributed training.
To illustrate, we consider two extreme cases.
The first case is that any two tensors are \textit{fully overlapped}, i.e., $\gamma_{G}^n = d_G$.
The left-hand term becomes $\log n$ while the right-hand term is $2(n-1)/n < 2$, which is much smaller. 
For typically large $n$, such as $n \ge 16$, the right-hand term is several times smaller.
The second case is that any two tensors have \textit{no overlaps}, i.e., $\gamma_{G}^k = k d_G$. 
The left-hand term becomes $n-1$, while the right-hand term is $(n-1)(1+1/n)$, which is slightly larger than $n-1$.
However, practical overlap ratios are much higher than 0.05, as shown in Figure~\ref{fig:overlap_ratio}.
More specifically, when $n = 16$, $\sum_{i=1}^{\log n} \gamma_G^{2^{i-1}} > \frac{n-1}{n} (\gamma_{G}^n+1)$ even if the overlap ratio of any two tensors is as low as 0.05.

\vspace{-0.05in}

\subsection{A Case Study on NMT model}
% \subsubsection{A Case Study on NMT}
% Numerical Comparison
\label{sec:numerical_comp}

% \begin{table*}[ht!]
% \vspace{0.03in}
% \footnotesize
% \centering
% \begin{tabular}{lllllll}
% \hline 
% Schemes          & Communication  & Aggregation  & Partition & Balance & Sparse format & Note \\ \hline  \hline                                      
% AGsparse~\cite{pytorch_ddp}         & Ring, Hierarchy, Point-to-point                   & One-shot           & Centralization    & N/A & COO           & \multirow{2}{*}{\shortstack{Cannot fully use overlaps \\ to reduce traffic.}} \\
% SparCML~\cite{sparcml}          & Hierarchy             & Incremental          & Centralization     & N/A & COO           &                                                                     \\ \hline      
% Sparse PS~\cite{PSosdi2014}        & Point-to-point        & One-shot             & Parallelism       & Imbalanced & COO           & \multirow{2}{*}{\shortstack{Suffer from imbalanced \\ communications.}}          \\
% OmniReduce~\cite{omnireduce}      & Point-to-point        & One-shot             & Parallelism       & Imbalanced & Tensor block  &                                                                     \\ \hline                               
% Zen              & Point-to-point        & One-shot             & Parallelism       & Balanced & Hash bitmap        &     This work. \\    
% \hline  
% \end{tabular}
% \vspace{-0.12in}
% \caption{Comparison of different synchronization schemes for sparse tensors based on their dimensions and data formats.}
% \label{table:comparison}
% \vspace{-0.18in}
% \end{table*}

In this section, We will empirically validate Theorem~\ref{thm:optimal_scheme}.
As listed in Table~\ref{table:comparison}, several synchronization schemes for sparse tensors are proposed~\cite{pytorch_ddp, sparcml, omnireduce}.
We will compare their performance from an algorithmic perspective, i.e., only consider their theoretical communication time and ignore other overheads, such as the computation time for aggregations and the sparse tensor encoding and decoding overheads.
% We only calculate the amount of traffic volume received by each GPU with each scheme without considering the system implementation.
% The analysis of the theoretical communication time of different schemes can be found in Appendix xxx.

% \noindent 
$\bullet$~\textbf{AGsparse~\cite{pytorch_ddp}.} It adopts [One-shot aggregation, Centralization], and separately collects non-zero gradients and the corresponding indices.
It cannot leverage the overlaps among sparse tensors to reduce the traffic volume. 
Note that there are different implementations for AGsparse with different communication patterns~\cite{MPIoptimization}.
% For example, let $C$ denote the overlap of all the sparse tensors, then $C$ will be received for $(n-1)$ times at each GPU, leading to unnecessary traffic volume.

% \noindent
$\bullet$~\textbf{SparCML~\cite{sparcml}.} It adopts {Hierarchical Centralization}.
It cannot leverage the overlaps among sparse tensors to reduce the traffic volume.
The performance of AGsparse and SparCML both depends on the overlaps.
The fewer overlaps, the closer their performance is to the optimal.
However, as shown in Figure~\ref{fig:overlap}, sparse tensors across GPUs in DL models have significant overlaps. 
% \textit{Note that SparCML is one of the optimal schemes  in Theorem~\ref{thm:optimal_scheme}.}

% \noindent
% \textbf{Sparse PS.}
% Parameter Servers (PS) architecture~\cite{PSosdi2014, kim2019parallax} is a synchronization scheme that adopts Point-to-point, Incremental aggregation, and Parallelism.
% It has two roles: workers and servers.
% For synchronizations of sparse tensors, workers push sparse tensors to servers. 
% After aggregation, workers pull aggregated tensors from servers to update model parameters.   
% We call this PS architecture for sparse tensors as \textit{Sparse PS} to distinguish it from the PS architecture for dense tensors.
% Because Sparse PS evenly partitions tensors, it suffers from imbalanced communications as discussed in Section~\ref{sec:performance_analysis}.

% \begin{figure}[t!]
%     \centering
% 	\includegraphics[width=1.0\linewidth]{figures/sparse_ps.pdf}
% 	\caption{An example of the load-imbalanced communications in Sparse PS. The numbers are the indices of the non-zero gradients. Most of the non-zero gradients from the three workers are pushed to Server 1, which becomes the communication bottleneck.}
% 	\label{fig:sparse_ps}
% \end{figure}

% \noindent
$\bullet$~\textbf{OmniReduce~\cite{omnireduce}.}
It adopts [Point-to-point, One-shot aggregation, and Parallelism].
OmniReduce consists of workers and aggregators.
It splits a gradient tensor into blocks of gradients and only sends non-zero blocks, i.e., blocks with at least one non-zero gradient, to aggregators for aggregations.
OmniReduce does not need to transmit indices for non-zero gradients.
It requires multiple aggregators for better scalability.
However, its performance suffers from imbalanced communications because it evenly partitions tensors.

% Now, let us consider a hypothetical scheme suggested by Theorem~\ref{thm:optimal_scheme} called Balanced Parallelism.

% $\bullet$~\textbf{Balanced Parallelism.}
% It has Point-to-point, Incremental aggregation, Parallelism, and Balanced communication.

The sparse data formats are as proposed by each scheme, i.e., AGsparse and SparCML use COO; OmniReduce uses tensor block.
We assume the sparse data format of Balanced Parallelism to be COO for a fair comparison.

\begin{figure}[t!]
    \centering
	\includegraphics[width=0.8\linewidth]{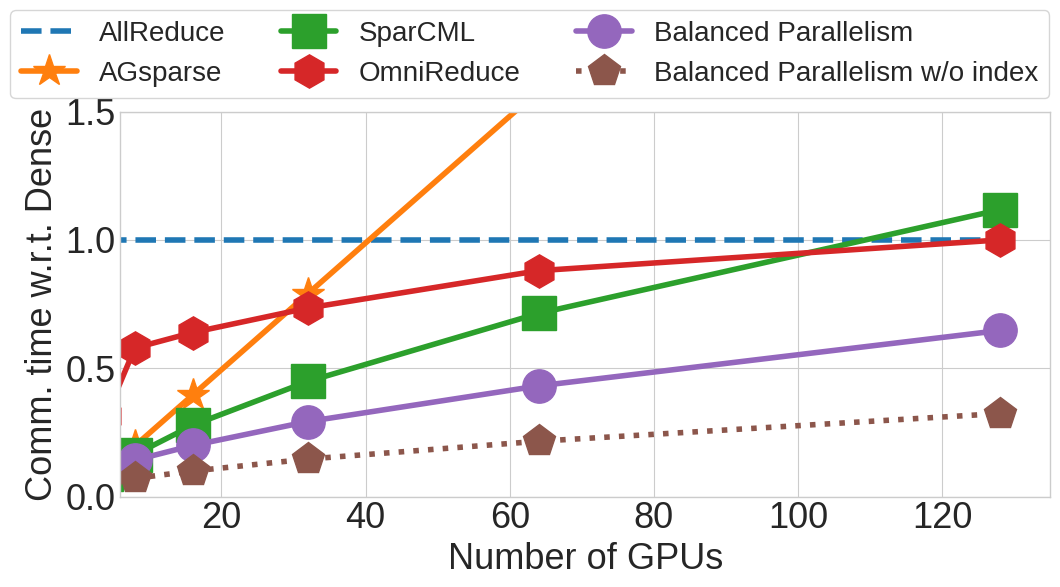}
	% \vskip -0.1in
	\caption{Comparison of different schemes for synchronizing sparse tensors in NMT~\cite{luong2015effective} with a batch size of 64.}
 % The sparse data formats are as each scheme proposed, i.e., OmniReduce uses tensor block; AGsparse and SparCML use COO. For a fair comparison, Balanced Parallelism uses COO.}
	\label{fig:the_time}
        \vspace{-0.2in}
\end{figure}

\vspace{3pt}
\noindent
\textbf{The case where Balanced Parallelism is optimal.}
Figure~\ref{fig:the_time} compares the performance of these synchronization schemes to synchronize sparse tensors in NMT with a batch size of 64. 
The comparison with other DL models has similar trends.
Their communication times are normalized to \textit{AllReduce}, which is the synchronization scheme for dense tensors.
The communication time of AGsparse linearly increases with the number of GPUs.
It performs worse than AllReduce with more than 40 GPUs because it does not leverage the overlaps among sparse tensors to reduce the traffic volume.
OmniReduce outperforms AllReduce with a small number of GPUs.
However, its performance improvement becomes marginal with more than 64 GPUs.
Due to the skewed distribution of non-zero gradients, most of the non-zero gradients are in one partition, leading to imbalanced communications.
In addition, tensors become denser after aggregation.
When splitting this partition into tensor blocks (e.g., each block has 256 gradients~\cite{omnireduce}), most of them are non-zero blocks.
Therefore, almost all gradients in this partition are sent to one aggregator and it becomes the communication bottleneck.
SparCML is worse than AllReduce with a large number of GPUs due to the duplicated indices and their gradients received at each GPU.
In contrast, Balanced Parallelism greatly outperforms existing synchronization schemes and AllReduce.
For example, existing schemes cannot reduce communication time compared to AllReduce with 128 GPUs, but the communication time of Balanced Parallelism is still 36\% lower than AllReduce.

\label{sec:comparision}

\vspace{3pt}
\noindent
\textbf{The case where SparCML is optimal.}
This case is rare in practical distributed training.
We have to reduce the batch size of NMT to a very small value for demonstration.
For example, when the batch size is 1 and the number of GPUs is 8, the density of the gradient tensors from the embedding layer is less than 0.1\%, and sparse tensors hardly have overlaps, i.e., $\gamma_G^k \approx k$.
In this scenario, SparCML outperforms AGsparse and Balanced Parallelism by 4\% and 9\%, respectively.
% The performance of OmniReduce significantly suffers from imbalanced communication and non-zero gradients in tensor blocks.
As the GPU number increases, SparCML still outperforms Balanced Parallelism, but the performance gap becomes more marginal.

\vspace{3pt}
\noindent
\textbf{The index communication overhead is non-negligible.}
When the sparse data format is COO, each non-zero gradient in a sparse tensor comes with an index, significantly inflating the traffic volume for gradient synchronization and communication time.
To demonstrate its costly overhead, Figure~\ref{fig:the_time} displays an ideal scheme named Balanced Parallelism without index, in which only non-zero gradients are synchronized with Balance Parallelism without any index information. 
Compared to the ideal scheme, Balanced Parallelism doubles the communication time.

\section{{\name} System}
\label{sec:zen}
 \vspace{-0.05in}

We crystalize the above observation and insights into a holistic gradient synchronization system, {\name}~(Figure~\ref{fig:overview}), which leverages the sparsity in DL models to minimize the synchronization time in distributed training. 
{\name} comprises both schemes described in Theorem~\ref{thm:optimal_scheme} to minimize communication time under different scenarios. 
At the runtime, \name~collects the lightweight sparsity profiling results of the first few iterations and intelligently determines the optimal scheme by comparing Equations~(\ref{eq:bp}) and~(\ref{eq:sm}).

% {\name} selects the optimal scheme by comparing Equations~(\ref{eq:bp}) and~(\ref{eq:sm}), or directly comparing $\frac{n-1}{n} (\gamma_{G}^n+1)$ with $\sum_{i=1}^{\log n} \gamma_G^{2^{i-1}}$.
% We observe that Balanced Parallelism is very likely the optimal synchronization scheme in practice.

% \vspace{-0.07in}
% \begin{observation}
% We have $\sum_{i=1}^{\log n} \gamma_G^{2^{i-1}} > \frac{n-1}{n} (\gamma_{G}^n+1)$ in practical distributed Deep Learning training.
% \end{observation}
% \vspace{-0.07in}

% We introduce two extreme cases as intuitions.
% The first case is that any two tensors have no overlaps and we have $\gamma_{G}^k = k d_G$.
% The left item becomes $n-1$ and the right item is $(n-1)(1+1/n)$, which is just slightly greater than $n-1$.
% The second case is that any two tensors are fully overlapped and we then have $\gamma_{G}^n = d_G$. 
% The left item becomes $\log n$ and the right item is $2(n-1)/n < 2$. 
% Because $n$ is typically large, e.g., $n \ge 16$, the right item is several times smaller than the left one.
% More specifically, when $n = 16$, $\sum_{i=1}^{\log n} \gamma_G^{2^{i-1}} > \frac{n-1}{n} (\gamma_{G}^n+1)$ even if the overlap ratio of any two tensors is only 0.05.
% However, the practical overlap ratio is much higher than 0.05, as shown in Figure~\ref{fig:overlap_ratio}.

We will focus on {Balanced Parallelism} in this section because no existing solution realizes it to date while SparCML~\cite{sparcml} has been proposed.
We first formulate the problem for Balanced Parallelism (\S\ref{sec:problem}) and then develop a hierarchical hashing algorithm to solve it (\S\ref{sec:hiera_hash}).
We then design a new sparse data format to minimize the communication overhead incurred by gradient indices (\S\ref{sec:hash_bitmap}) in sparse tensors.

% \vspace{-0.15in}
% \subsection{Balancing Communications}
% \label{sec:load_balance}

% \vspace{-0.05in}
% \vspace{-0.1in}
% \label{sec:problem}

\vspace{-0.15in}
\subsection{Balanced Parallelism Formulation}
%\subsection{\name's~Theoretical Foundation}
\label{sec:problem}
\vspace{-0.03in}

% Balanced Parallelism has the same communication pattern, aggregation pattern, and partition pattern as Sparse PS, but its communications are always well-balanced.
For convenience, we borrow the concepts of workers and servers from Parameter Servers (PS) architecture~\cite{PSosdi2014} to Balanced Parallelism.
Because there are two steps in Balanced Parallelism for gradient synchronization, we also call them Push and Pull, respectively.

Suppose there are $n$ workers and $n$ servers in Balanced Parallelism. 
$I_i \subset \mathbb{N_+} $ is the indices of non-zero gradients generated by worker $i$.
We define the problem of achieving balanced parallelism as follows:

\vspace{-0.02in}
\begin{problem}\label{prob:load_balance} 
Let $I$ denote the union of $\{I_1, I_2, \cdots, I_n\}$. 
We would like to have a mapping $f:I\to [n]$ such that:
\begin{enumerate}[noitemsep,nolistsep,leftmargin=*] 
    \item For every $i\in [n]$ and $j\in [n]$, the cardinality of set $\{idx \in I_i|f(idx)=j\}$ is equal to $|I_i|/n$.
    \item For every $j\in [n]$, the cardinality of set $\{idx \in I|f(idx)=j\}$ is equal to $|I|/n$.
\end{enumerate}
\end{problem}
\vspace{-0.02in}

Here we elaborate more on the two requirements for the mapping $f$ accordingly as below:
\vspace{2pt}
\begin{enumerate}[noitemsep,nolistsep,leftmargin=*] 
    \item \textbf{Load balance in Push.} For every worker, mapping $f$ needs to decompose its non-zero gradients evenly into $n$ partitions. Therefore, workers can transmit the same amount of non-zero gradients to each server.
    \item \textbf{Load balance in Pull.} {Each of the servers should have the same number of non-zero gradients after aggregation.} It also implies that the same index from different workers should be sent to the same server.
\end{enumerate}
% \noindent Problem~\ref{prob:load_balance} assumes the same number of workers and servers, but it is easy to generalize this problem to cases in which the number of workers and servers are different.

% \vspace{-0.1in}
% \subsubsection{Strawman Solutions}
% \label{sec:strawman}
% \vspace{-0.07in}

\vspace{-0.02in}
\begin{definition}[The imbalance ratio]
\label{def:imbalance_ratio}
Given a mapping $f$ that decompose $I_i$ into $n$ partitions, denoted as $\{I_i^1, \cdots, I_i^n\}$. The imbalance ratio of Push with $f$ is $\max_{i, j \in [n]} \{n|I_i^j|/|I_i| \}$.

Let $I$ denote the union of $\{I_1, I_2, \cdots, I_n\}$ and the sets of indices at the $n$ servers after aggregation are $\{\mathbb{I}_1, \mathbb{I}_2, \cdots, \mathbb{I}_n\}$.
The imbalance ratio of Pull with $f$ is $\max_{i \in [n]} \{n|\mathbb{I}_i|/|I| \}$.
\end{definition}
\vspace{-0.02in}

{Based on Definition~\ref{def:imbalance_ratio}, the imbalance ratio of Push and Pull in Balanced Parallelism is 1. }
Our goal is to minimize the imbalance ratio for any distributions of non-zero gradients.

\vspace{2pt}
\noindent 
\textbf{Data-dependent solutions cause costly overheads.} 
Due to different sets of indices on different workers, data-dependent solutions need to analyze their overall distribution and calculate one mapping for all workers, inevitably incurring non-negligible computation overheads.
In our testbed, the measured computation cost is orders of magnitude greater than the iteration time.
Hence, we cannot afford to apply a data-dependent algorithm and obtain a mapping $f$ for every iteration. 
A possible approach is to compute $f$ periodically and maintain it for the next iterations. 
However, this approach still leads to high imbalance ratios due to the varying index distributions across iterations. 
One strawman following this approach is to sort the index set $I$, evenly partition it into $n$ parts, and use the boundary indices as the thresholds to partition the index sets in the next iterations. 
When computing the thresholds every 1000 iteration for NMT model~\cite{luong2015effective} with $n=16$ and applying these thresholds to the following iterations, the imbalance ratio of push fluctuates between $1.4$ and $5.1$, causing imbalanced communications among servers.
Moreover, the imbalanced communications introduced by data-dependent solutions make it hard to estimate the iteration time. 
Many resource scheduling mechanisms for GPU clusters assume predictable and stable iteration times for allocating resources to training jobs~\cite{narayanan2020gavel, qiao2021pollux, peng2018optimus, mahajan2020themis}.
It is cumbersome to schedule GPU resources with fluctuating communication time.

 \vspace{-0.16in}
\subsection{Data-independent Hierarchical Hashing}
\label{sec:hiera_hash}
 \vspace{-0.03in}

Due to the limitations of data-dependent solutions, we must develop a data-independent solution to achieve load balance in both Push and Pull with negligible computation overheads.

% Hashing algorithms have been widely applied to address load imbalance problems in various domains, such as distributed system~\cite{karger1997consistent, stoica2001chord} and distributed database~\cite{ellis1983extendible, litwin1996lh}.
% However, it is challenging to apply existing hashing algorithms to address Problem~\ref{prob:load_balance} because 1) they can lead to information loss of gradients and harm training accuracy, and 2) they can cause costly computation overheads to perform hash functions.

% \noindent 
% \textbf{Naive hashing algorithms on GPUs are lossy.} 
A naive solution to address Problem~\ref{prob:load_balance} is to apply a universal hash function across multiple threads on a GPU, i.e., each thread independently operates hash functions on a disjoint input and writes them into a hash memory (pseudocode in Appendix~\ref{app:straw_alg}).
Unfortunately, this approach is lossy.
When two indices are hashed to the same location in a hash memory, only one index can be written into the hash memory, while the other is overwritten, leading to significant information loss.
For example, when the hash memory size equals the size of a dense tensor with 10\% density, 17.5\% gradients are lost due to hash collision.
Increasing the hash memory size can reduce the information loss, but it introduces non-negligible computational overhead because the algorithm must extract the non-zero values from the hash memory after the hash operations.
To illustrate, consider the embedding layer (1.6GB) in LSTM~\cite{LSTM}.
When the hash memory size is four times the tensor size, the information loss rate decreases to 4.8\%. 
However, the extraction overhead increases to 41.8ms using the built-in \texttt{nonzero()} API in PyTorch 2.2 on an NVIDIA V100 GPU.
This overhead is unacceptable, as it is roughly 40\% of the single-GPU iteration time (114 ms in our testbed). 
In addition, increasing the hash memory size significantly increases GPU memory usage, potentially causing out-of-memory issues and crashing the training process.

We develop a novel hierarchical algorithm to solve Problem~\ref{prob:load_balance}.
We leverage multiple threads in GPUs to efficiently perform hash functions; we achieve no information loss and minimal GPU memory usage by using four techniques.

\vspace{2pt}
\noindent
\textbf{Technique \#1: communication-oriented hash memory management.} 
The hash memory is divided into multiple partitions and the data written into each partition is for the corresponding server.
Each partition is further divided into parallel memory and serial memory to prevent data loss. 
When a thread within a GPU executes a hash function, it initially examines for hash collisions by checking if the hashed location is occupied.
In the absence of a collision, the data is written to the parallel memory, enabling concurrent writing operations across all threads. 
In case of a collision, the colliding indices are sequentially written to the serial memory using an atomic operation, allowing only one thread to write data at a time.
Although the indices written in the memory are in a random order, there is no need to sort them because their orders do not affect the aggregated results.

Unfortunately, we observe that the expense of serial writing becomes significant when the hash collision rate is elevated.

\noindent
\textbf{Technique \#2: multiple hash functions for each thread in GPU.} 
Multiple hash functions are used to reduce the cost of serial writing even given a small hash memory size.
When a hash collision occurs, a GPU thread can rehash the index with a new hash function to another location in the parallel memory.
There is a chance that this new location is available and the number of sequential write operations to the serial memory can be reduced.
Although hash collision still exists even with multiple hash functions, we observe that the collision rate is less than 1\% with four hash functions and the cost of serial writing into the serial memory becomes acceptable.

However, rehashing with multiple hash functions can cause incomplete aggregations.
Because different GPUs have different sets of indices for non-zero gradients, their sequences of indices being hashed are also different.
Therefore, the location of a particular index can be different across GPUs.
For example, two indices $idx_1$ and $idx_2$, where $idx_1 < idx_2$, are hashed to the same location with the first hash function.
GPU 1 has $idx_2$; GPU 2 has both $idx_1$ and $idx_2$.
In GPU 1, the location of $idx_2$ is determined by the first hash function, but in GPU 2, the location of $idx_2$ is determined by the second hash function because the location hashed by the first one has been occupied by $idx_1$. 
Subsequently, partitioning the memory will lead to the same index assigned to different partitions at different GPUs, resulting in incomplete aggregations.

\noindent
\textbf{Technique \#3: consistent hierarchical hashing across workers.} 
We propose a two-level hierarchical hashing algorithm to guarantee complete aggregations.
The first-level hashing determines the partition that an index belongs to and guarantees that an index will belong to the same partition across all GPUs.
The second-level hashing determines its locations in this partition. 
To ensure that the same index from different workers can be sent to the same server, {\name} allocates the same first-level hashing function to all the workers, but allows them to have independent second-level hashing functions.

\vspace{-0.06in}
\noindent
\textbf{Technique \#4: lock-free read-after-write mechanism.} 
There is a concern that two values from two threads can be hashed to the same available location simultaneously and this hash collision cannot be detected by conventional mechanisms, leading to information loss.
{\name} uses a read-after-write mechanism to check this collision, eliminating this type of information loss.
% The probability of this special case becomes negligible with the probability less than $10^{-4}$ in our implementation on GPUs. 
% This probability is even lower than the floating point rounding error on GPUs~\cite{hillesland2004gpu,whitehead2011precision}. 
After memory writing, a thread reads the value stored in the location and this operation has no dependency on the values from other threads.
If the value equals what it writes, this thread will move to the next input.
However, if the value is not what it writes, it implies there is a hash collision and its value is overwritten. 
Then the thread will take a rehash or serial writing.

\begin{figure}[t!]
\vspace{0.05in}
\begin{center}
\includegraphics[width=0.9\linewidth]{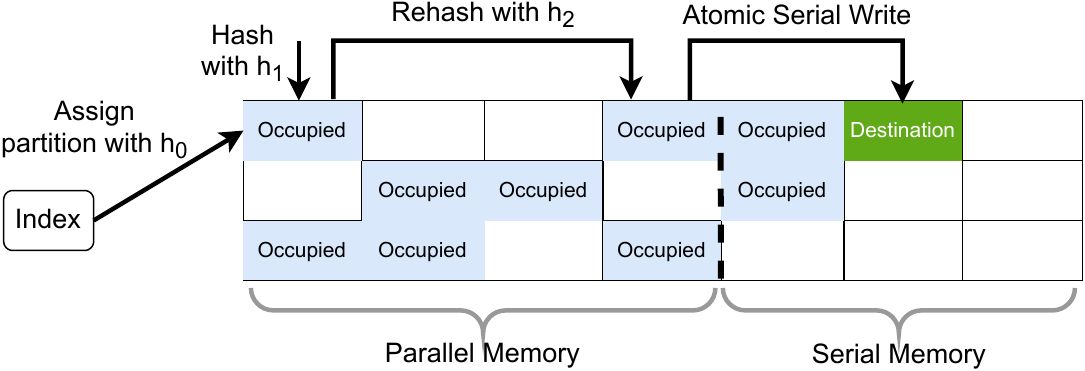}
% \vskip -0.05in
\caption{Demonstration of the hierarchical hashing algorithm. We perform parallel hashing on the indices. For each index, we use hash function $h_0$ to assign its partition. We next use hash function $h_1$ to assign it to the first location. However, because this location is occupied, we rehash it with function $h_2$ to the fourth location. As it is also occupied, we serially write the index into the serial memory with an atomic operation.}
\vspace{-0.16in}
\label{fig:rehash}
\end{center}
\end{figure}

\begin{algorithm}[t!]
\footnotesize
\caption{Hierarchical Hashing Algorithm}
\label{alg:hash}
\SetKwInput{KwInput}{Input}                % Set the Input
\SetKwInput{KwOutput}{Output}              % set the Output
\DontPrintSemicolon 
  
    \KwInput{ $G$ is a dense tensor and $I \subset \mathbb{N}_{+}$ is the indices of its non-zero gradients. $n$ is the number of partitions. Each partition has a memory size $r_1+r_2$, where $r_1$ and $r_2$ are the memory sizes for parallel and serial operations, respectively. $h_0:\mathbb{N}_{+}\to [n]$ is a universal hash function. $H=\{h_1, \cdots, h_{k}\}$ is a set of  universal hash functions where $h_i:\mathbb{N}_{+}\to [r]$. \;
    }
    \KwOutput{The partitioned sparse tensors.}
    
    % Set Function Names
    \SetKwFunction{FMain}{hierarchical\_hash}

    \SetKwProg{Fn}{Function}{:}{}
    \Fn{\FMain{$I$, $G$, $h_0$, $H$}}{
        Allocate memory $x\gets \mathbf{0}^{n\times (r_1+r_2)}$ \; 
        % Allocate memory $x_{\mathsf{serial}}\gets \mathbf{0}^{n\times r_1}$ \; 
        Allocate atomic counters $c \gets \mathbf{r_1}^n$ \;
        % Generate universal hash function $h_0:\mathbf{N}\to [n]$ \;
        % \For{$j\gets1$ \KwTo $k$}{
        %     Generate universal hash function $h_j:\mathbf{N}\to [r_1]$ \;
        % }
        \ForEach{$idx \in I$ {\rm{\textbf{in parallel}}}}{
            $p\gets h_0(idx)$\;
            
            \For{$i\gets1$ \KwTo $k+1$}{
                $q\gets h_i(idx)$\;
                \If{$i = k+1$} {
                    $q \gets \texttt{atomicAdd}(c[p], 1)$\;
                    $x[p][q] \gets idx $\;
                }
                \If{$x[p][q] == 0$}
                {         $x[p][q]\gets idx $\;
                \textbf{break} \;
                }
            }
        
        }
        $output = []$\;
        \For{$i\gets 0$ \KwTo $n-1$ }{
            $indices = \texttt{nonzero}(x[i])$\;
            $values = G[indices]$\;
            $output\text{.append}((indices, values))$\;
        }
        \KwRet $output$
    }
\end{algorithm}

\setlength{\textfloatsep}{0.5cm}
\setlength{\floatsep}{0.5cm}

\vspace{-0.07in}
\noindent
\textbf{Hierarchical hashing algorithm.}
The pseudocode with the four techniques is shown in Algorithm~\ref{alg:hash}.
An example of this algorithm is also illustrated in Figure~\ref{fig:rehash}.
Given a dense tensor $G$ and the indices of its non-zero gradients $I$,
it allocates a memory $x$ with shape $n\times (r_1+r_2)$, where $n$ is the number of partitions, $r_1$ is the memory size for parallel hashing operations, and $r_2$ is the serial memory size. 
It performs a hashing operation for every $idx \in I$ in parallel (Lines 4-17).
A universal hash function $h_0:\mathbb{N}_{+} \to [n]$ is used to locate $idx$ to partition $p = h_0(idx)$ (Line 5). 
The algorithm also needs $k$ universal hash functions $H = \{h_1, \cdots, h_k\}$ with $h_i: \mathbb{N}_{+} \to [r_1]$.
After determining the partition $p$, the algorithm attempts to find an available destination $x[p][h_1(idx)]$ with $h_1$.
If this location is available, $idx$ is written into it.
Otherwise, the algorithm rehashes $idx$ with $h_2$ to find a new location.
It rehashes an index for at most $k$ rounds and uses $h_i$ as the hash function for round $i$ until it finds an available destination (Lines 6-16).
The algorithm writes $idx$ to the serial memory of partition $p$ after $k$ rehash attempts fail
%The algorithm writes it to the serial memory allocated to partition $p$ after $k$ rehashes
(Lines 8-11).
Serial writing is an atomic operation (Lines 9-10) to ensure no information loss.
Once all indices are written into the memory, it extracts sparse tensors from the memory (Lines 19-23).

\noindent
\textbf{Negligible extraction overhead.} 
Thanks to multiple hashing functions and the serial memory, Algorithm~\ref{alg:hash} can achieve no information loss with a small memory size.
The incurred overhead to extract the indices from the memory after hashing (Line 20 in Algorithm~\ref{alg:hash}) becomes negligible. 

% Next, we will analyze the imbalance ratio of Algorithm~\ref{alg:hash}.

\noindent
\textbf{Guaranteed imbalance ratio.}
Because the hash function $h_0$ determines the partition of each index, the imbalance ratio of Algorithm~\ref{alg:hash} is guaranteed by the following theorem.

\vspace{-0.05in}
\begin{theorem}[Load Balance of Algorithm~\ref{alg:hash}]
\label{thm:load_balance:informal}
Given a dense tensor $G$ with $|G|$ parameters.
Algorithm~\ref{alg:hash} provides a mapping $f:I \to [n]$ such that
\begin{enumerate}[noitemsep,nolistsep,leftmargin=*] 
    \item With probability at least $1-1/n$, its imbalance ratio of Push is at most $1 + \Theta(\sqrt{\frac{n \log n}{|G|d_G}}) $. 
    \item With probability at least $1-1/n$, the imbalance ratio of Pull is at most 
    $1 + \Theta(\sqrt{\frac{n \log n}{|G|d_G^n}}) $.
\end{enumerate}
\end{theorem}
\vspace{-0.1in}

\noindent
The proof can be found in Appendix~\ref{app:proof_th2}. 
Because $n \log n$ is orders of magnitude smaller than $|G|$, Algorithm~\ref{alg:hash} performs a good approximation to the exact solution of Problem~\ref{prob:load_balance} for both Push and Pull.
For example, suppose $n=128$, $|G|=10^7$, and $d_G^n = 0.5$, $\sqrt{\frac{n \log n}{|G|d_G^n}} < 0.02$.
Its practical imbalance ratio is always less than 1.1 for the six studied models.

\noindent
\textbf{No dependency on workloads.}
Note that we impose no assumptions on the data distributions and only use the property of universal hashing algorithms. 
Hence, Algorithm~\ref{alg:hash} can obtain a general theoretical guarantee for different distributions of non-zero gradients in DL training workloads. 
  
% {\name} can support different optimizer, such as SGD~\cite{sgd}, Adam~\cite{projectadam}, and AdaGrad~\cite{duchi2011adaptive}.
% The computation of an optimizer has two steps: gradient synchronization and parameter update.
% {\name} decouples the two steps following BytePS~\cite{byteps_osdi}.
% It aims to minimize the communication time of gradient synchronization.
% After sparse tensors are synchronized, GPUs have the same aggregated gradients, with which they can update the parameters of the Deep Learning model individually.

% \begin{figure}[t!]
% \centering
% \begin{minipage}{.48\linewidth}
%   \flushleft
%   \includegraphics[width=\linewidth]{figures/serial_write_perc.png}
%   % \vskip -0.1in
%   \captionof{figure}{The percentage of serial writing versus the number of hashes.}
%   % \vspace{-0.1in}
%   \label{fig:serial_write}
% \end{minipage}
% \hspace{\fill}%
% \begin{minipage}{.49\linewidth}
%   \flushright
%   \includegraphics[width=\linewidth]{figures/bitmap.pdf}
%   \vskip -0.13in
%   \captionof{figure}{Bitmap still incurs costly overhead to represent indices in Zen.}
% %   \vskip -0.1in
%   \label{fig:bitmap}
% \end{minipage}
% \end{figure}

\vspace{-0.1in}
% \subsection{Minimizing the Indices Overhead}
% \subsection{Cost-Efficient BitMap Indexing}
\subsection{Cost-Efficient Encoding Scheme}
\label{sec:hash_bitmap}
\vspace{-0.03in}
\subsubsection{Existing Sparse Formats are Inefficient}
\vspace{-0.03in}

There are several sparse formats to represent sparse tensors for their synchronization.
Unfortunately, none of them can minimize the overhead incurred by the indices for non-zero gradients.
We assume the data type of gradients is FP32.

% \noindent
$\bullet$~\textbf{COO.}
It is efficient with a low tensor density~\cite{wang2022dragonn, dgc}.
However, it doubles the traffic volume and becomes inefficient for a high density.
As shown in Figure~\ref{fig:denser_ratio}, tensors get denser after aggregation.
For example, the average tensor density of BERT increases from 1.06\% to 40.8\% after the aggregation of sparse tensors from 128 GPUs.
Theoretically, transmitting sparse tensors in Pull can reduce the traffic volume by $2.5\times$ compared to transmitting dense tensors, but the reduction shrinks to $1.2\times$ due to the indices for non-zero gradients.

% \begin{figure}[t!]
%     \centering
% 	\begin{subfigure}[t]{0.6\linewidth}
% 	\includegraphics[width=\linewidth]{figures/bitmap_even.pdf}
% 	\vspace{-0.15in}
% 	\caption{Tensors are evenly partitioned.}
% 	\label{fig:bitmap_even}
%     \end{subfigure}  
%     \begin{subfigure}[t]{0.6\linewidth}
% 	\includegraphics[width=\linewidth]{figures/bitmap_hash.pdf}
% 	\vspace{-0.15in}
% 	\caption{The non-zero gradients are randomly distributed after Algorithm~\ref{alg:hash} is applied. }
% 	\label{fig:bitmap_hash}
%     \end{subfigure} 
%     \vskip -0.1in
%     \caption{Bitmap works when tensors are evenly partitioned, but it incurs costly overhead in Load-balanced PS since non-zero gradients in each server are randomly distributed.}
%     \label{fig:bitmap}
% \end{figure}

% \noindent
$\bullet$~\textbf{Tensor block.}
It is used in OmniReduce~\cite{omnireduce}. 
A dense tensor is split into blocks of gradients and only non-zero blocks are transmitted.
However, it is also inefficient when the tensor density is high.
When splitting a tensor with high density into tensor blocks (e.g., each block has 256 gradients), most of them have at least one non-zero gradient and become non-zero tensor blocks.
The synchronization scheme has to transit almost all the gradients.
% Moreover, non-zero gradients after aggregation at each server are randomly distributed.
% It likely that different servers have the same non-zero blocks, leading to incomplete aggregations.
% For example, the gradient with Index 3 is aggregated at Server 0 and the gradient with Index 5 is aggregated at Server 1.
% The first block in both Server 0 and Server 1 are non-zero block.
% Therefore, workers have to receive the first block twice from both servers and they have to aggregate the first block again for the final aggregated results.

\begin{figure}[t!]
\vspace{0.05in}
\centering
  \includegraphics[width=0.65\linewidth]{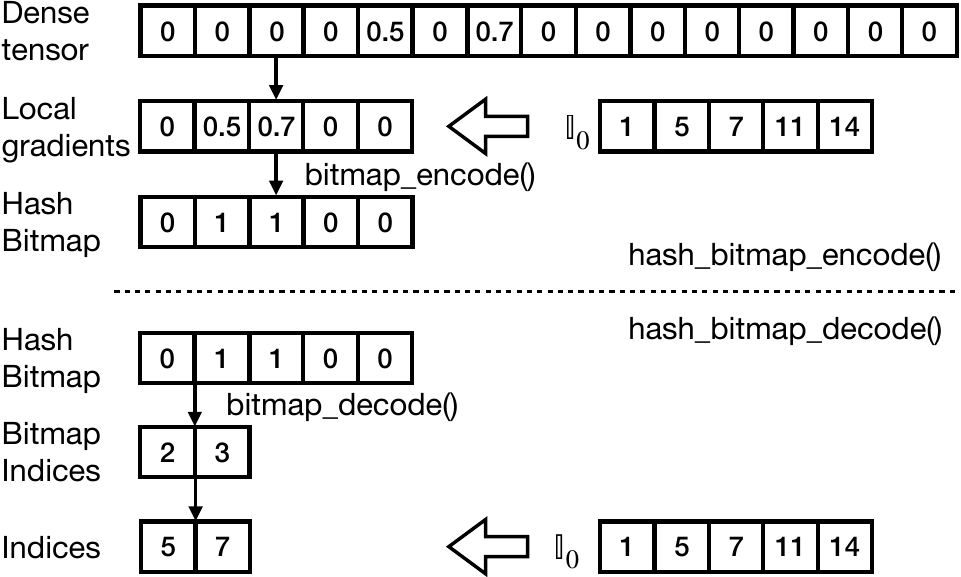}
  \vspace{-0.07in}
  \captionof{figure}{An illustration of the hash bitmap.}
  \label{fig:hash_bitmap}
  \vspace{-0.1in}
\end{figure}

% \noindent
$\bullet$~\textbf{Bitmap.}
It only needs one bit to indicate whether a gradient is zero or not.
Unfortunately, a conventional bitmap still incurs non-negligible traffic volume to identify non-zero gradients.
When the dense tensor $G$ is evenly partitioned, the indices of non-zero gradients in each server are in a sub-range of $[1, |G|]$, where $|G|$ is the number of gradients in $G$.
For example, when $|G| = 15$ and there are three servers, the index range in Server $i$ is $[5i + 1, 5(i+1)]$. 
The extra bitmap size required to represent the indices of non-zero gradients in each server is $|G|/n/32$ when the data type of gradients is FP32. 
The total bitmap size received by each worker is constantly $|G|/32$.
When the tensor size of $G$ is 856MB, which equals the embedding table size in DeepFM, the total bitmap size is 27 MB.
Although Algorithm~\ref{alg:hash} enables load balance, the non-zero gradients in each server are randomly distributed in the whole range.
If we still use a bitmap to represent the indices, the extra bitmap size in each server is $|G|/32$ and the total bitmap size received at each worker becomes $n|G|/32$, which linearly increases with the number of servers.
When there are 16 servers, the total bitmap size is 428MB.

\vspace{-0.15in}
\subsubsection{Hash Bitmap}
\vspace{-0.03in}

We develop a novel \textit{hash bitmap} for {\name} to minimize the overhead to represent indices for non-zero gradients in Pull.
Given a dense tensor $G$ and $h_0$ in Algorithm~\ref{alg:hash}, the set of indices $\mathbb{I}_i = \{idx \in [1, |G|] ~| ~h_0(idx) = i\}$ in each worker that should be pushed to Server $i$ is determined, though it is not in a continuous range.
Since $\mathbb{I}_i$ and $\mathbb{I}_j$ are disjoint when $i \ne j$, it provides an opportunity to construct the bitmap based on $\mathbb{I}_i$, rather than the whole range.

Figure~\ref{fig:hash_bitmap} illustrates how the hash bitmap works for $\mathbb{I}_0$ with $|G| = 15$ and three servers.
The indices for the two non-zero gradients are $\{5, 7\}$.
\texttt{hash\_bitmap\_encode()} is used to encode the indices.
Given a dense tensor $G$, it first extracts the local gradients according to the indices in $\mathbb{I}_0$.
It then encodes the local gradients into a bitmap. 
Because the second and the third gradients are non-zero, the second bit and the third bit in the hash bitmap are 1 and the other bits are 0.
\texttt{hash\_bitmap\_decode()} is used to decode the hash bitmap to a set of indices.
It first decodes a hash bitmap to the bitmap indices, which are the indices of 1.
For example, because the second and the third bits in the hash bitmap are 1, the bitmap indices are $\{2, 3\}$.
It then uses the bitmap indices as the indices to extract the corresponding values in $\mathbb{I}_0$ as the global indices for non-zero gradients.
In this example, the values are $\{5, 7\}$, which are exactly the indices for the two non-zero gradients.
The pseudocode is shown in Algorithm~\ref{alg:hash_bitmap}.

\begin{algorithm}[t] \small
% \vspace{0.05in}
\footnotesize
\caption{The hash bitmap}
\label{alg:hash_bitmap}
\SetKwInput{KwInput}{Input}                % Set the Input
\SetKwInput{KwOutput}{Output}              % set the Output
\DontPrintSemicolon 
    % Set Function Names
    \SetKwFunction{FEncode}{hash\_bitmap\_encode}
    \SetKwFunction{FDecode}{hash\_bitmap\_decode}
    \KwInput{$G$ is a dense tensor. $\mathbb{I}_i = \{idx \in [1, |G|] ~| ~h_0(idx) = i\}$, where $h_0$ is defined in Algorithm~\ref{alg:hash}.}
    \SetKwProg{Fn}{Function}{:}{}
    
    \Fn{\FEncode{$G$, $\mathbb{I}_i$}}{
        $local\_gradients = G[\mathbb{I}_i]$ \;
        $hash\_bitmap = \texttt{bitmap\_encode}(local\_gradients)$ \;
        \KwRet $hash\_bitmap$ \;
    }

    \Fn{\FDecode{$\mathbb{I}_i$, $hash\_bitmap$}}{
        $bitmap\_indices = \texttt{bitmap\_decode(hash\_bitmap)}$ \;
        $indices = \mathbb{I}_i[bitmap\_indices]$ \;
        \KwRet $indices$ \;
    }
\end{algorithm}

The function \texttt{hash\_bitmap\_encode()} is invoked at each server, which then broadcasts the hash bitmap to all the workers.
After each worker receives the hash bitmaps from all the servers, it invokes \texttt{hash\_bitmap\_decode()} to decode the hash bitmaps to the indices with the corresponding $\mathbb{I}_i$.
Note that $\mathbb{I}_i$ is computed and sorted offline and it remains unchanged for the same $h_0$ in both servers and workers.

The hash bitmap guarantees that the total hash bitmap size received at each worker from all servers is constantly $|G|/32$ in Pull of {\name}.
The explanation can be found in Appendix~\ref{app:proof_th3}. 
Zen still uses COO to represent sparse tensors in Push due to the low tensor density. 
% The benefit of replacing COO with hash bitmap is limited. 

\begin{figure*}[t!]
\vspace{0.05in}
    \centering
	\begin{subfigure}[t]{0.33\linewidth}
	\includegraphics[width=0.95\linewidth]{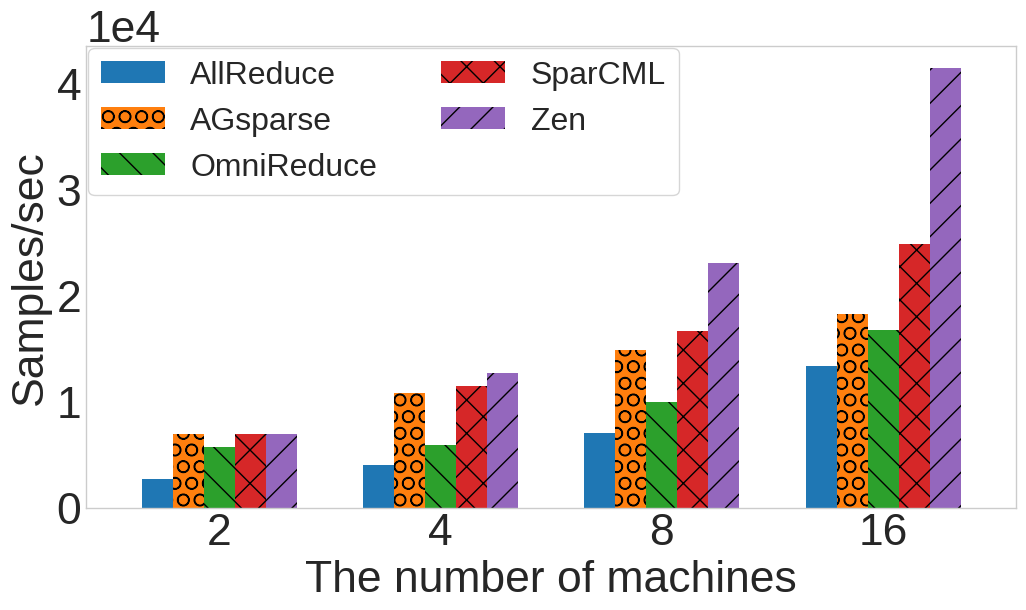}
	\vspace{-0.08in}
	\caption{LSTM}
	\label{fig:lstm_25G}
    \end{subfigure}  
    \begin{subfigure}[t]{0.33\linewidth}
	\includegraphics[width=0.95\linewidth]{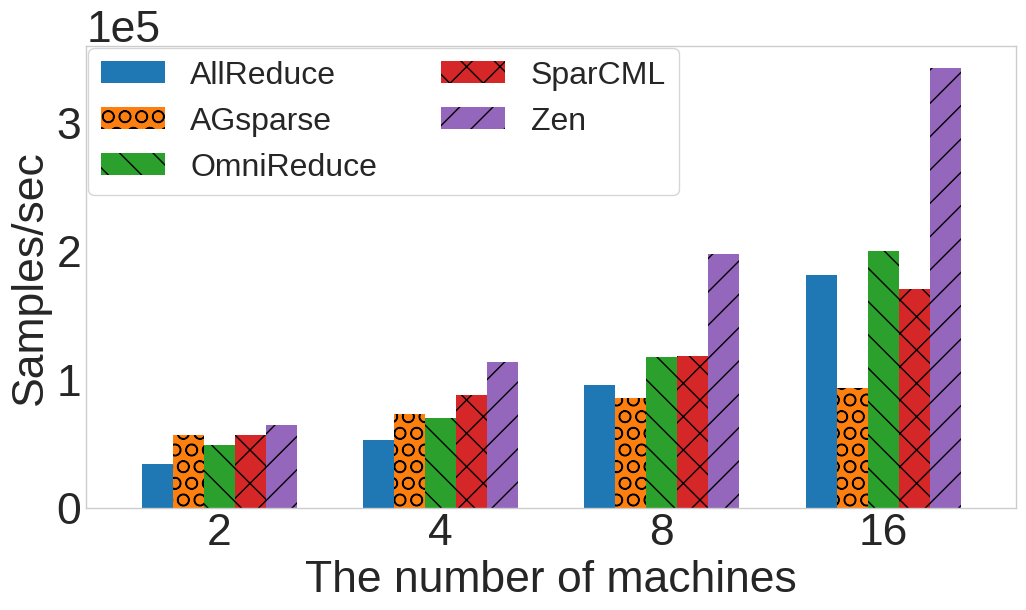}
	\vspace{-0.08in}
	\caption{DeepFM}
	\label{fig:deepfm_25G}
    \end{subfigure} 
    \begin{subfigure}[t]{0.33\linewidth}
	\includegraphics[width=0.95\linewidth]{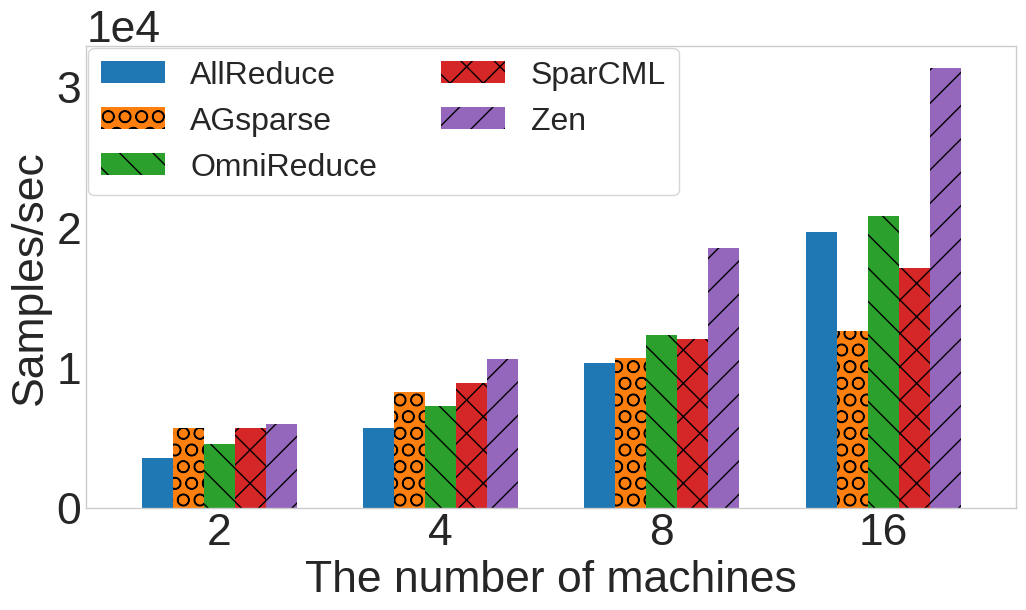}
	\vspace{-0.08in}
	\caption{NMT}
	\label{fig:nmt_25G}
    \end{subfigure}  
    \vspace{-0.15in}
    \caption{Training throughput of models with natural tensor sparsity.}
    \vspace{-0.15in}
    \label{fig:25G}
\end{figure*}

\begin{figure*}[t!]
    \centering
	\begin{subfigure}[t]{0.33\linewidth}
	\includegraphics[width=0.95\linewidth]{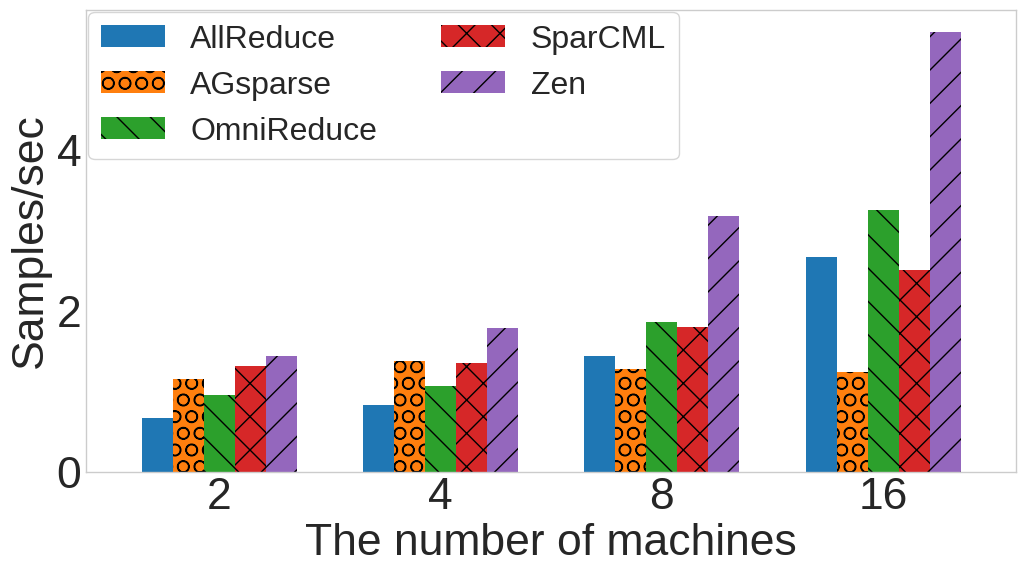}
	\vspace{-0.08in}
	\caption{Llama3.2-3B}
	\label{fig:llama_25G}
    \end{subfigure}  
    \begin{subfigure}[t]{0.33\linewidth}
	\includegraphics[width=0.95\linewidth]{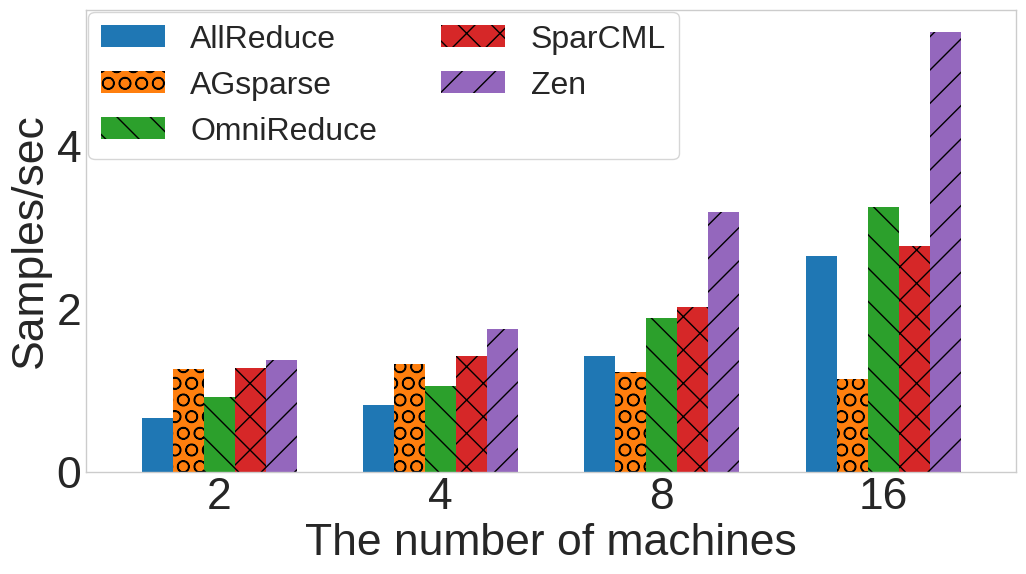}
	\vspace{-0.08in}
	\caption{OPT2.7B}
	\label{fig:opt_25G}
    \end{subfigure} 
    \begin{subfigure}[t]{0.33\linewidth}
	\includegraphics[width=0.95\linewidth]{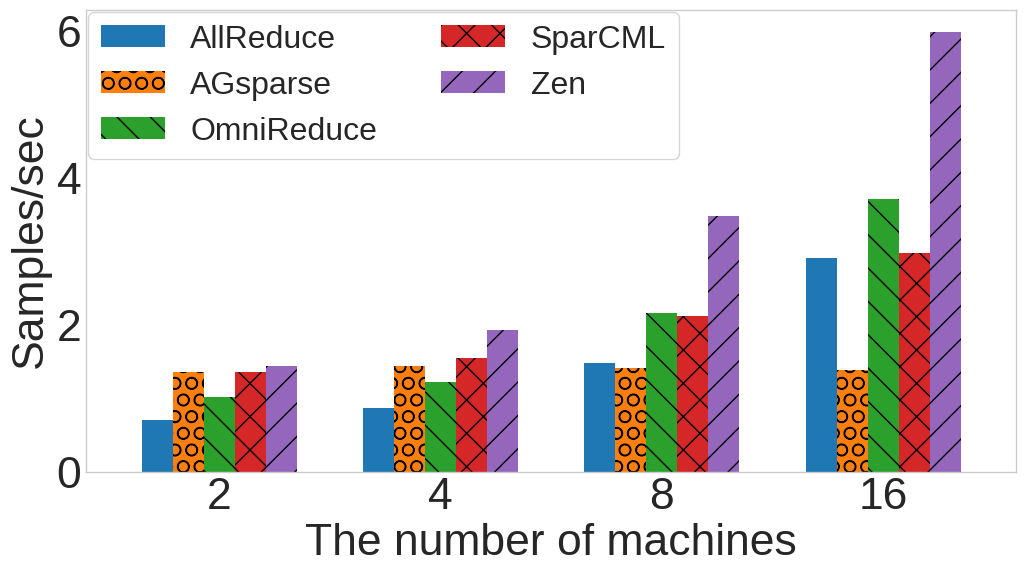}
	\vspace{-0.08in}
	\caption{Gemma2-2B}
	\label{fig:gemma_25G}
    \end{subfigure} 
    \vspace{-0.15in}
    \caption{Training throughput of models with tensor sparsity from sparsification algorithms.}
    \vspace{-0.15in}
    \label{fig:100G}
\end{figure*}

% \vspace{-0.1in}
\subsection{Implementation}
\vspace{-0.03in}

{We implement \name~with about 900 lines of Python code and 250 lines of CUDA code, with another 500 lines of code for hacking ColossalAI~\cite{li2023colossal}.}
% \textcolor{blue}
For DL model training, we use the ColossalAI to implement data parallelism and hybrid parallelism (data parallelism + tensor Parallelism, with tensor parallelism degree set to 8, typically within a single node). 
ColossalAI, based on our testing, outperforms DeepSpeed~\cite{rasley2020deepspeed} and Megatron-LM~\cite{megatronlm}, while also being user-friendly for deployment. 
Additionally, we extend ColossalAI to support the Gemma2~\cite{team2024gemma} model from Google with custom policies for efficient tensor parallelism. 

For natural tensor sparsity, we only apply sparse tensor synchronization schemes to gradients from the embedding layers for their communications across machines.
For the sparsification algorithm, we use DGC~\cite{dgc} and adapt it for tensor parallelism: sampling data locally on each device, gathering these samples across all devices, computing a global top-k threshold based on the aggregated data, and applying this threshold locally to determine top-k values on each device. 
We register a \texttt{custom\_comm\_hook} in PyTorch's DDP~\cite{pytorch_ddp} to enable DGC and our synchronization scheme.
Gradient computations are overlapped with gradient synchronization during backward propagation.
The gradient tensors are fused into communication buckets in DDP before applying DGC to tensors~\cite{wang2023cupcake}.
We determined a 128MB bucket size that works best for the evaluated models.

We implement the hierarchical hash algorithm in CUDA C and use it as an extension for PyTorch.
The used hash function is MurmurHash~\cite{appleby2008murmurhash}. 
We set different seeds for MurmurHash to generate different hash functions. 
At the beginning of training, {\name} generates a set of random seeds and broadcasts them to all GPUs to ensure hash consistency among workers.
{\name} communicates dense tensors within machines with ReduceScatter/AllGather~\cite{gropp2002mpich2, MPIoptimization} because NVLink is typically equipped in GPU machines.
In our evaluations, Balanced Parallelism is the optimal scheme for the studied models.

%-------------------------------------------------------------------------------

%-------------------------------------------------------------------------------
% \input{implementation}
%-------------------------------------------------------------------------------

%-------------------------------------------------------------------------------
\vspace{-0.15in}
\section{Evaluation}
\label{sec:evaluation}
\vspace{-0.1in}

% \vspace{-0.03in}
\subsection{Experimental Setup}
\label{exp_setup}
% \vspace{-0.05in}
\noindent \textbf{Testbeds.} 
We use 16 p3.16xlarge instances from AWS EC2 for our evaluations.
Each machine has 8 NVIDIA V100 GPUs (16GB GPU memory) and NVLink is equipped to support intra-instance GPU-to-GPU communications.
The CPU memory size is 488GB and all the instances are connected by a 25Gbps network.
Each machine runs Ubuntu 20.04 and the software environment includes CUDA-11.0, PyTorch-2.2, NCCL-2.7.8, CuPy-11.0, and ColossalAI-v0.4.6.

% \noindent \textbf{Workloads.}
% \Zhuang{add LLM description}
% \begin{table}[h!]
% \footnotesize
% \centering
% \scalebox{0.9}{
% \begin{tabular}{ccccc}
% \hline
% \textbf{Model size} & \textbf{Hidden size} & \textbf{Intermediate} & \textbf{\#Layers} & \textbf{\#AH} \\
% \hline
% OPT 2.7B\cite{opt175b} & 2560 & 10240 & 32 & 32 \\
% LLAMA3.2 3B\cite{dubey2024llama} & 3072 & 8192 & 28  & 24 \\
% GEMMA2 2B\cite{team2024gemma} & 2304 & 9216 & 26 & 8 \\
% \hline
% \end{tabular}}
% \caption{Configurations of different language models. AH is short for attention heads. OPT 2.7B means OPT with 2.7 billion parameters. The same naming convention applies to other models.}
% \label{tab:model_configs}
% \end{table}

\noindent \textbf{Workloads.} 
For natural tensor sparsity evaluation, we use three models, LSTM~\cite{LSTM}, DeepFM~\cite{guo2017deepfm}, and NMT~\cite{luong2015effective}, as listed in Table~\ref{tab:model_sparsity}.
The used datasets are One Billion Word~\cite{1b-word}, Criteo~\cite{criteo}, and IWSLT 2014 De-En~\cite{iwslt}, respectively.
The used batch sizes are 128, 1024, and 64, respectively. 
These models can fit into the memory of each GPU.
The per-GPU batch size is kept constant as the number of GPUs increases.
For tensor sparsity from sparsification algorithms, we use three LLMs, Llama3.2-3B, OPT-2.7B, and Gemma2-2B, as listed in Table~\ref{tab:llm_configs}.
We use RedPajama~\cite{weber2024redpajama} as the training corpus. 
Because these models cannot fit into a single GPU, we use tensor parallelism (TP) with degree 8.
The batch size within a TP group is 1 and the sequence length is 1024 tokens, ensuring efficient training without exceeding GPU memory limits.

\noindent \textbf{Baselines.}
We compare {\name} with AGsparse~\cite{pytorch_ddp}, SparCML (SSAR\_Recursive\_double)~\cite{sparcml}, and OmniReduce~\cite{omnireduce}.
% The SparCML paper uses CPUs for aggregations of sparse tensors and it incurs prohibitive computation overheads~\cite{sparcml}. For a fair comparison, we have implemented it with GPUs for aggregations.}
We use \texttt{AllReduce}~\cite{horovod} for the synchronization of dense tensors.

\vspace{-0.1in}
\subsection{DL Training Evaluation}
\label{sec:e2e_exp}
\vspace{-0.03in}

In this section, we present the training efficiency of {\name} on the six models and compare it with the baselines. 
We set $k=3$, $r_1 = 2|G|d_G$, and $r_2 = r_1/10$ for Algorithm~\ref{alg:hash}. 
% We perform training on $2$, $4$, $8$ and $16$ GPU machines with both $25$Gbps and $100$Gbps networks.

\noindent
\textbf{Training throughput improvement.}
Figure~\ref{fig:25G} shows the training throughput of models with natural tensor sparsity.
{\name} outperforms all baselines by processing more samples in a second.
When training LSTM with 16 machines, {\name} achieves up to a $1.67 \times$ speedup over SparCML (the best baseline), a $2.48 \times$ speedup over OmniReduce, and a $3.1 \times$ speedup over AllReduce. 
In both DeepFM and NMT, the best baseline is OmniReduce.
{\name} achieves $1.44\times$ speedup and $1.51\times$ speedup over OmniReduce for DeepFM and NMT, respectively.
As the number of machines increases, the benefits of {\name} over SparCML and OmniReduce are enlarged, indicating {\name}'s great scalability.
{\name} still has great speedups for the gradient compression scenario, as shown in Figure~\ref{fig:100G}.
For Llama3.2-3B, it achieves up to a $1.68 \times$ speedup over OmniReduce, a $2.19 \times$ speedup over SparCML, and a $2.02 \times$ speedup over AllReduce. 
In OPT2.7B and Gemme2-2B, {\name} achieves up to $2.10 \times$ and $2.04 \times$ speedups over AllReduce, and $1.66 \times$ and $1.61 \times$ speedups over OmniReduce.
% Similar to the $25$Gbps setting, {\name} also approaches the upper bound with different numbers of GPUs.
These results demonstrate that {\name} fully leverages sparsity in DL models to optimize training efficiency.
\begin{figure}[t] \small
\centering
    \includegraphics[width=0.7\linewidth]{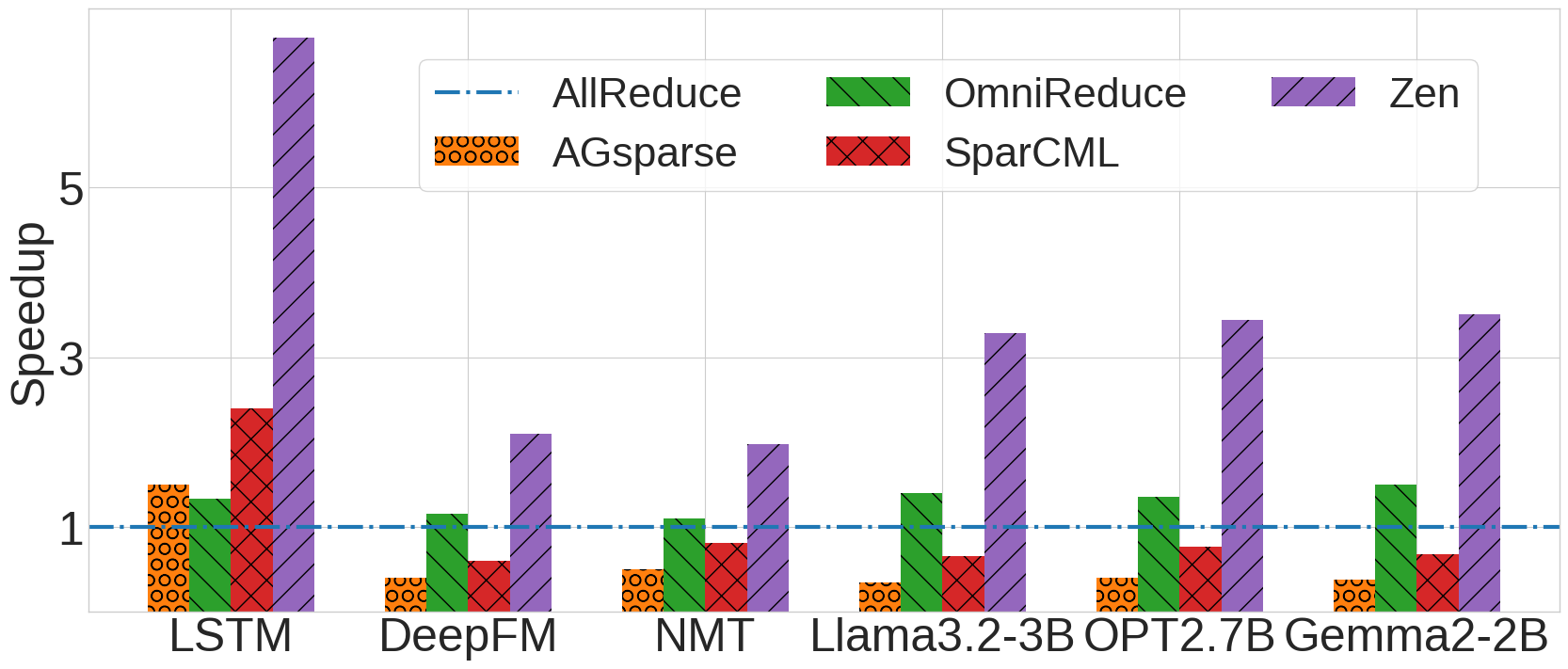}
    \vspace{-0.12in}
    \caption{Communication speedups over AllReduce.}
      \label{fig:comm_speedup}
    \vspace{-5pt}
\end{figure} 

\noindent
\textbf{Communication improvement.} 
The performance of {\name} is driven by the reduction in communication time.
Figure~\ref{fig:comm_speedup} shows the speedups of different synchronization schemes over AllReduce with 16 machines.
The speedup of OmniReduce is up to $1.58\times$.
We observe the performance of AGsparse, Sparse PS, and SparCML can be even worse than AllReduce in some cases.
They use COO as the sparse format.
With a high density, the sparse tensor size with COO is larger than the dense tensor size.
The communication time of AGsparse linearly increases with the number of machines.
With SparCML, the overlaps among sparse tensors can be received multiple times at each GPU.
In contrast, {\name} achieves $6.77\times$ speedup for LSTM and $3.51\times$ speedup for Gemma2-2B.
Its speedups over SparCML and OmniReduce are up to $2.82\times$ and $5.16\times$, respectively.
{\name} also achieves $2.10\times$ speedup for DeepFM and $3.44\times$ speedup for OPT2.7B over Allreduce.

\vspace{-0.1in}
\subsection{Understanding \name}\label{sec:exp_micro}
\vspace{-0.05in}

\begin{figure}[t!]
    \centering
	\begin{subfigure}[t]{0.47\linewidth}
	\includegraphics[width=0.95\linewidth]{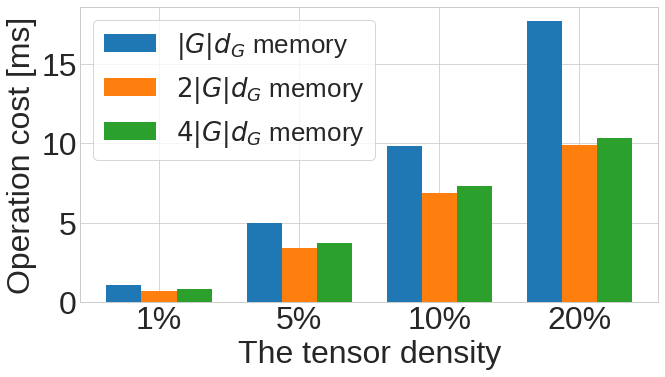}
	\vspace{-0.1in}
	\caption{}
	\label{fig:memory_cost}
    \end{subfigure}  
    \begin{subfigure}[t]{0.47\linewidth}
	\includegraphics[width=0.95\linewidth]{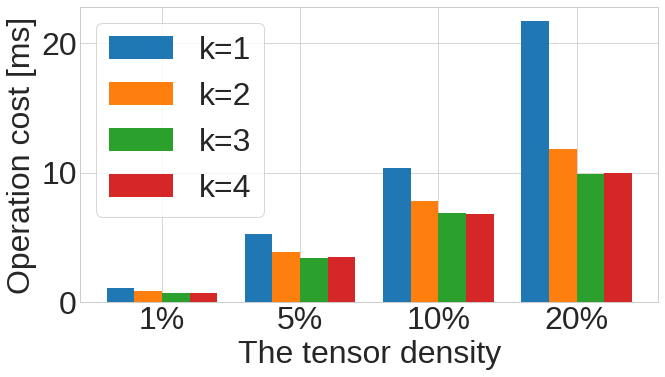}
	\vspace{-0.1in}
	\caption{}
	\label{fig:rehash_cost}
    \end{subfigure} 
    \vspace{-0.1in}
    \caption{The computation overhead of Algorithm~\ref{alg:hash}. (a) Different memory sizes and (b) Different numbers of rehash.}
    \label{fig:algo2_cost}
\end{figure}

\noindent
\textbf{A study on parameters for Algorithm~\ref{alg:hash}.} 
We simulate a tensor with $214$M parameters (same as the embedding gradients in DeepFM) and vary its density to perform a study on both $r_1$ and $k$ in Algorithm~\ref{alg:hash}.
We first study parameter $r_1$. 
We set $r_2 = r_1/10$ and $k=3$. 
As shown in Figure~\ref{fig:memory_cost}, when we increase $r_1$ from $|G|d_G$ to $2|G|d_G$, there is a notable reduction in the operation cost because the larger memory size increases the probability of successful parallel writing and reduces the workload of serial writing. 
But when we further increase $r_1$ from $2|G|d_G$ to $4|G|d_G$, it leads to a higher computation time. 
There are two reasons behind this effect. 
Firstly, the workload of serial writing is already low with $2|G|d_G$ memory. 
Further increasing $r_1$ only marginally helps reduce the computation overhead. 
Secondly, a larger memory size can increase the computation overhead to extract the indices (see Algorithm~\ref{alg:hash}) and thus degrade the overall performance.
Figure~\ref{fig:rehash_cost} shows the computation costs versus $k$ when we use $2|G|d_G$ memory. 
Increasing $k$ from $1$ to $3$ can help reduce the operation cost as it alleviates the serial writing workload, but $k=3$ and $k=4$ have very similar operation costs.

\begin{figure}[t]
  \subfloat[\label{fig:hash_bitmap_cmp}]{\includegraphics[width=0.47\linewidth]{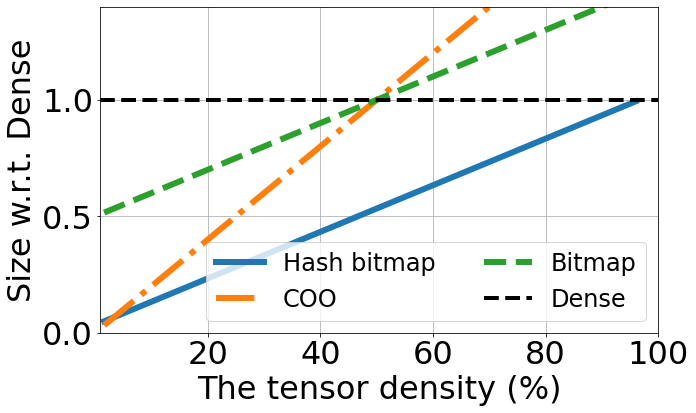}}
  \subfloat[\label{fig:hash_bitmap_speedup}]{\includegraphics[width=0.51\linewidth]{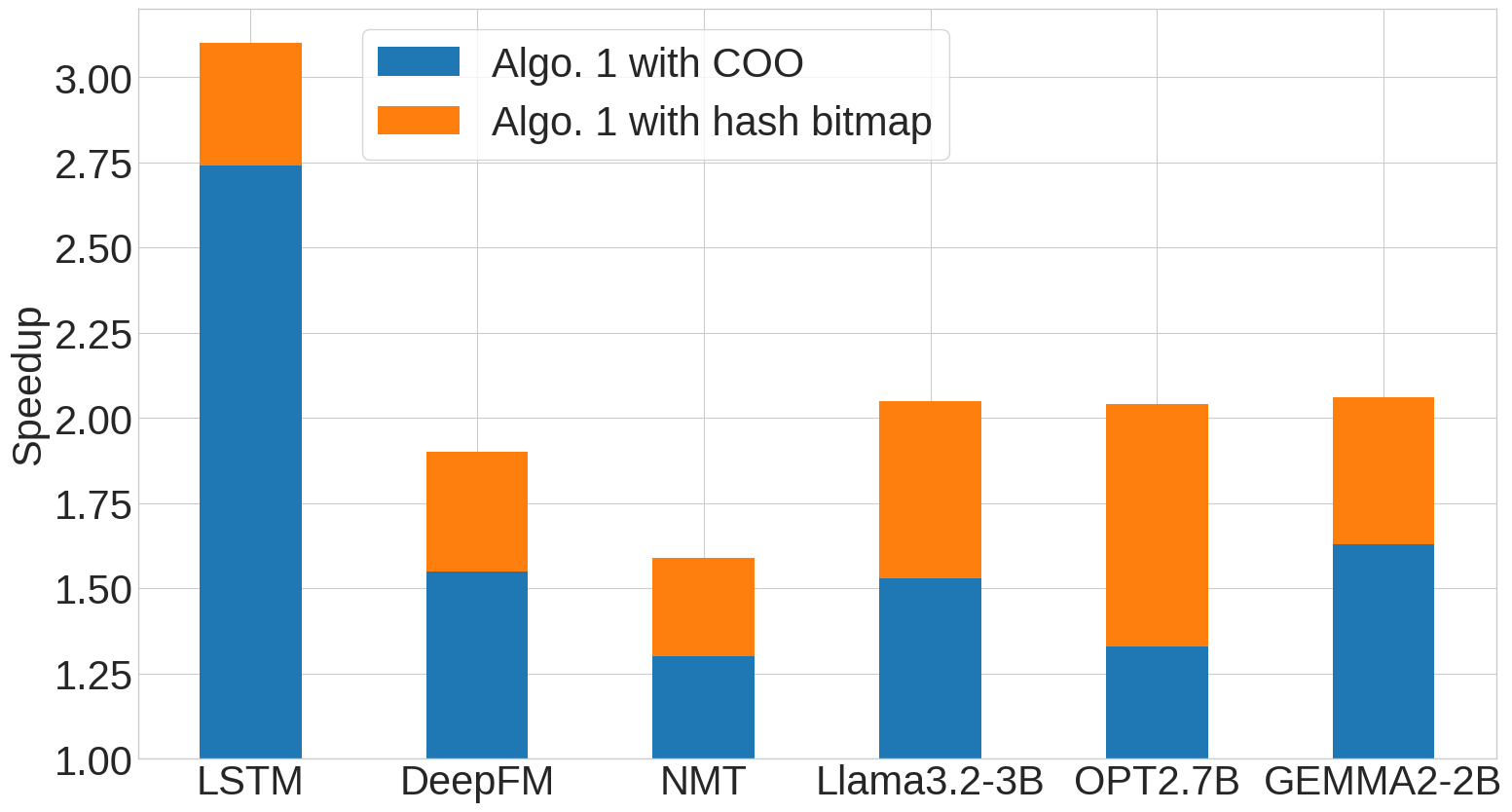}}
  \vspace{-0.1in}
\caption{(a) The effectiveness of the hash bitmap; (b) The performance breakdown of {\name}. }
\end{figure}

\noindent
\textbf{Hash Bitmap.}
We show the effectiveness of the hash bitmap in representing the indices of non-zero gradients. 
Figure~\ref{fig:hash_bitmap_cmp} shows the tensor data size with different sparse formats.
The sizes are normalized to the dense tensor and there are 16 servers.
The tensor density is the total density of all servers after aggregation.
The gap between the hash bitmap and COO increases with the tensor density.
It also significantly outperforms the bitmap. 
Besides, the hash bitmap can still reduce the traffic volume with a density of 95\% compared to the dense tensor, but the bitmap and COO cannot save the volume when the density is greater than 50\%.
The performance of tensor blocks varies with the distribution of non-zero gradients. 
Some sparse tensors in the studied DL models can even transmit higher traffic volume than COO because a non-zero block has more zero gradients than non-zero gradients.

\noindent
\textbf{{Performance Breakdown.}}
We break down the performance of {\name} by Algorithm~\ref{alg:hash} and the hash bitmap. 
Figure~\ref{fig:hash_bitmap_speedup} illustrates the training throughput speedup breakdown over AllReduce with 16 machines.
It can be seen that the primary performance benefits of {\name} come from Algorithm~\ref{alg:hash}, with the hash bitmap format providing noticeable additional benefits.
For example, when the data format is COO after applying Algorithm~\ref{alg:hash}, the speedup is $2.74\times$ and $1.53\times$ for LSTM and Llama3.2-3B, respectively.
Replacing COO with the hash bitmap can further improve the speedups by 13\% and 34\%.

\vspace{-0.2in}
\section{Related Work}
\label{sec:related_work}
\vspace{-0.12in}
Related work on schemes to support sparse tensors synchronization has already been discussed in Section~\ref{sec:numerical_comp}.

\vspace{2pt}
\noindent
\textbf{Acceleration of dense tensor synchronization.}
% Many works are proposed to accelerate synchronizations of dense tensors.
ATP~\cite{atp} and SwitchML~\cite{SwitchML} exploit programmable switches for the synchronization of dense tensors.
BytePS~\cite{byteps_osdi} uses spare CPU and bandwidth resources in GPU clouds to optimize communications.
These approaches disregard the values of gradients and communicate all gradients.
In contrast, Zen leverages sparsity in DNN models and only transmits non-zero gradients to reduce the synchronization time.

\vspace{2pt}
\noindent
\textbf{Acceleration of sparse tensor synchronization.}
Parallax~\cite{kim2019parallax} utilizes Sparse PS and it cannot achieve balanced communications in synchronization.
{\name} can achieve balanced communications by using a novel hierarchical hashing algorithm.
% EmbRace~\cite{li2022embrace} optimizes communications of sparse tensors in model parallelism, but {\name} focuses on data parallelism.
Flare~\cite{de2021flare} and Libra~\cite{pan2022libra} use programmable switches to accelerate sparse tensor communications, but they rely on specific hardware. 
% \textcolor{red}
{Ok-Topk~\cite{li2022near} proposes a novel sparse allreduce algorithm; however, its inherently data-dependent balancing strategy results in significant overhead.}
% Moreover, 
{\name} analyzes the characteristics of sparse tensors and explores the design space for communication schemes to determine the optimal one, but prior approaches did not consider these factors.

\vspace{2pt}
\noindent
\textbf{Hash algorithms for load balance.} 
Previous works attempt to achieve load balance using hashing~\cite{azar1994balanced,cao2000performance,czumaj2003perfectly,berenbrink2006balanced,wieder2017hashing}.
They typically assign two partitions to a given index using two hash functions and then select the partition with more available memory~\cite{mitzenmacher2001power,richa2001power,dahlgaard2016power, mitzenmacher2017probability}. 
% This line of works utilizes the power of two choices~\cite{mitzenmacher2001power,richa2001power,dahlgaard2016power, mitzenmacher2017probability} in hashing.
However, these methods require serial writing of indices to memory and cannot leverage multiple threads in GPUs. 
In contrast, {\name} enables parallelizable computing on GPUs.
While DRAGONN~\cite{wang2022dragonn} introduces a hash-based algorithm for parallel writing of non-zero gradients to memory, it does not address the imbalanced communications in distributed training and cannot handle hash collisions, resulting in information loss.
In contrast, {\name} achieves balanced communications and avoids information loss.

\vspace{-0.2in}
\section{Conclusion}
\label{sec:conclusion}
\vspace{-0.1in}

We make two primary contributions in this work. First, we systematically explore the design space to identify the optimal synchronization schemes for sparse tensors in distributed training.
Second, we propose {\name}, a practical design to pursue the optimal synchronization schemes with a novel hashing algorithm using parallel computing on GPUs without information loss, that greatly improves the efficiency of sparse DNN model training without any impact on model accuracy.
% Our realization of this scheme, {\name}, is data-independent and includes a novel hashing algorithm that can use parallel computing on GPUs to achieve well-balanced communications among GPUs without information loss. 

%-------------------------------------------------------------------------------

\bibliographystyle{plain}
\bibliography{reference}
% \clearpage
\clearpage

\appendix
\section*{Appendix}

\section{Algorithm for a Strawman Solution}\label{app:straw_alg}

In this section, we present Algorithm~\ref{alg:strawman_hash}, which is the algorithm for the data-independent solution with a straightforward hashing algorithm.
This algorithm supplements the discussions in Section~\ref{sec:hiera_hash}.

\begin{algorithm}[ht!]
\footnotesize
\caption{A strawman solution with hashing}
\label{alg:strawman_hash}
\SetKwInput{KwInput}{Input}                % Set the Input
\SetKwInput{KwOutput}{Output}              % set the Output
\DontPrintSemicolon 
  
    \KwInput{ $G$ is a dense tensor and $I \subset \mathbb{N}_{+}$ is a set of indices of its non-zero gradients. $n\in \mathbb{N}_{+}$ is the number of partitions. $r\in \mathbb{N}_{+}$ is the memory size for each partition. $h:\mathbb{N}_{+}\to [nr]$ is an universal hash function. \;
    }
    \KwOutput{The partitioned sparse tensors.}
    
    % Set Function Names
    \SetKwFunction{FMain}{Main}
 
    \SetKwProg{Fn}{Function}{:}{}
    \Fn{\FMain{$I$, $G$, $h(\cdot)$}}{
        Allocate memory $x\gets \mathbf{0}^{n\times r}$ \; 
        % \tcp{Could be parallelized in $s_{\mathsf{index}}$}
        \ForEach{$idx \in I$ {\rm{\textbf{in parallel}}}}{
            $p\gets \lfloor h(idx)/r \rfloor$\;
            $q\gets  h(idx) \mod r$\;
            $x[p][q]\gets idx$\;
        }
        $output = []$\;
        \For{$i\gets 0$ \KwTo $n-1$}{
            $indices = \texttt{nonzero}(x[i])$\;
            $values = G[indices]$\;
            $output.append( (indices, values))$\;
        }
        
        \KwRet $output$;
    }
\end{algorithm}

Given a dense tensor $G$, the set of indices of its non-zero gradients is $I$. 
The algorithm first allocates a memory $x$ with shape $n\times r$, where $n$ represents the number of partitions and $r$ is the memory size for each partition. 
For every $idx \in I$, it uses a given universal hash function~\cite{carter1977universal} $h:\mathbb{N}_{+} \to [nr]$ to generate the hash value $h(idx)$, where $nr$ is the range of hash function $h$.
Next, it writes the $idx$ to the $(h(idx)\mod r)$th location in partition $\lfloor h(idx)/r \rfloor$.
The hashing operation is performed in parallel to minimize the computation overhead~\cite{wang2022dragonn}.
After that, it extracts the non-zero indices from the memory of each partition and uses them to look up the corresponding gradients from $G$. 
Finally, it returns a sparse tensor for each partition and pushes them to the corresponding servers.

\section{Theoretical Analysis}\label{app:proof}

\subsection{Proof of Theorem 1}
\label{app:proof_th1}

\begin{customthm}{1}[Optimal schemes]
To minimize communication time for sparse tensors, the optimal synchronization scheme is either {Balanced Parallelism} or {Hierarchical Centralization}.
\end{customthm}

We prove Theorem~\ref{thm:optimal_scheme} with two lemmas. 

\begin{customlemma}{1}
When the partition pattern is fixed to Parallelism, the optimal scheme is  {Balanced Parallelism}.
\end{customlemma}

\begin{proof}
There are three communication patterns: Ring, Hierarchy, and Point-to-point. 
We first consider synchronization schemes with Point-to-point communication and Parallelism, namely, the PS architecture.
Given a gradient tensor $G$ with the density of $d_G$ and there are $n$ servers.
We first analyze the communication time of push and pull operations separately.
We then discuss the communication time of different PS schemes.

\noindent
\textbf{Push.}
Because the skewness ratio is $s_G^n$, the largest density in the $n$ partitions is $s_G^n d_G$.
The size of the sparse tensor extracted from this partition is $2 s_G^n d_G M/n$. 
As a result, the communication time of push in sparse PS is $2 (n-1)s_G^n d_G M/n/B$.

\noindent
\textbf{Pull.}
After aggregation, the largest density in the $n$ partitions becomes $s_G^n d_G^n$.
In existing implementations of the PS architecture, the communication time of Pull is $2 (n-1)s_G^n d_G^n M/n/B$ because each server needs to broadcast its aggregated results to all the workers with point-to-point communications~\cite{byteps_osdi, PSosdi2014}.
In theory, there are other ways to implement Pull in the PS architecture.
For example, each server can perform a \texttt{broadcast} collective operation.
The performance of \texttt{broadcast} with different algorithms is analyzed in~\cite{MPI_vs_NCCL} and its communication time for Pull can be expressed as 
$2 b d_G^n M/B$, where $b$ is the number of rounds in an algorithm. 
For example, $b=\lceil \log n \rceil$ when it uses Binomial Tree Algorithm and $b = \frac{2(n-1)}{n}$ when it uses Scatter-AllGather Algorithm~\cite{gropp2002mpich2, MPI_vs_NCCL}.

\noindent
\textbf{Sparse PS.}
Combining the communication time of push and pull with point-to-point communications, 
it overall communication time is $2(n-1)(d_G+d_G^n)s_G^n M/n/B$. 

\noindent
\textbf{Sparse PS with the broadcast.} 
When considering \texttt{broadcast} for Pull, the overall communication time becomes $2 (n-1)s_G^n d_G M/n/B + 2 b d_G^n M/B$.
We denote this case as sparse PS with the broadcast.

\noindent
\textbf{Balanced Parallelism.}
In Balanced Parallelism, the skewness ratio $s_G^n$ is always 1.
We replace the $s_G^n$ in the communication time of sparse PS as 1 and have the communication time for Balanced Parallelism: $2(n-1)(d_G+d_G^n) M/n/B$.

\noindent
\textbf{Balanced Parallelism is optimal among schemes with Parallelism.}
It is clear that Balanced Parallelism is much better than sparse PS when the skewness ratio is large.
The performance ratio of PS with broadcast to Balanced Parallelism is $\frac{s_G^n}{1+\gamma_G^n} + \frac{n}{n-1}\frac{b \gamma_G^n}{1 + \gamma_G^n} > \frac{s_G^n + b \gamma_G^n}{1 + \gamma_G^n} $. Because both $s_G^n$ and $b$ are greater than 1, the ratio is also greater than 1.
Hence, Balanced Parallelism always outperforms sparse PS and sparse PS with broadcast in terms of communication time.

Because One-shot aggregation cannot leverage the overlaps among sparse tensors, 
the performance of synchronization schemes with One-shot aggregation is worse than those with Incremental aggregation.
Therefore, we only consider Incremental aggregation for schemes with Ring communication or Hierarchy communication.
We consider the best case for them, i.e., the skewness ratio is 1 after tensor partition with Parallelism.
In addition, we only need to compare the first step because they have the same communication time in the second step.
The communication time of the first step in Balanced Parallelism is $2(n-1)d_GM/n/B$.

\noindent
\textbf{Schemes with Ring and Incremental Aggregation.} 
They have $n-1$ communication stages. 
The tensor density in the $i_{th}$ stage is $d_G^{i}$. Note that $d_G^{1} = d_G$.
Therefore, the communication time is $2\sum_{i=1}^{n-1} d_G^i M/n/B$.
Because tensors can get denser after aggregation, we have $d_G^i \le d_G^j$ when $i < j$ and $\sum_{i=1}^{n-1} d_G^i \ge (n-1)d_G$.
As a result, the communication time of schemes with Ring and Incremental aggregation is no less than that of Balanced Parallelism.

\noindent
\textbf{Schemes with Hierarchy and Incremental aggregation.} 
They have $\log n$ communication stages.
Because each partition has a hierarchical structure, the total traffic volume in the $i_{th}$ stage is $ \frac{d_G^{2^{i-1}}}{2^{i-1}} Mn$ and the total traffic volume in all the $\log n$ stages is $V = \sum_{i=1}^{\log n} \frac{d_G^{2^{i-1}}}{2^{i-1}} Mn$. 
Because $d_G^{2^{i-1}} \ge d_G$, we have $V \ge \sum_{i=1}^{\log n} \frac{d_G}{2^{i-1}} Mn = 2(n-1)d_GM$.
Therefore, the traffic volume received at each GPU is no less than $2(n-1)d_GM/n$ and the communication time is no less than that of Balanced Parallelism.
\end{proof}

\begin{customlemma}{2}
When the partition pattern is fixed to Centralization, the optimal scheme is Hierarchical Centralization.
\end{customlemma}

\begin{proof}
When any two sparse tensors have no overlaps, the minimum traffic volume each GPU has to receive is all the tensors from other GPUs.
Any synchronization scheme with Centralization achieves the optimal communication time.

When sparse tensors are overlapped, let $n$ denote the number of GPUs and $I_0, I_1, \dots, I_{n-1}$ are the set of indices for non-zero gradients in each GPU, respectively.
$C$ is the overlap of all the sparse tensors, i.e., $C = \bigcap I_i$.
If a synchronization scheme adopts point-to-point communication or one-shot aggregation, each GPU has to receive $C$ for $n-1$ times.
Then we consider Incremental aggregation and the communication pattern is Ring or Hierarchy.
With Ring, the tensor from each GPU is aggregated at each stage and then forwarded to the next GPU.
Consequently, this tensor is received by every GPU.
Because $C$ is the common overlap, each GPU also has to receive $C$ for $n-1$ times.
When the communication pattern is Hierarchy, each GPU receives data from all the other GPUs with its own hierarchical structure that has $\log n + 1$ levels, as shown in Figure~\ref{fig:hierarchy}.
Because the root GPU is in each level, it has to receive $C$ for $\log n$ times.
It suggests that the traffic volume in the scheme with Hierarchy, Incremental aggregation, and Centralization is less than that in other schemes with Centralization.
Let $C'$ denote the overlap of a subset of the sparse tensors, we can have a similar conclusion that a subset of GPUs has to receive $C'$ multiple times.
In other words, each GPU still has to receive the overlaps multiple times.
\end{proof}

Lemma~\ref{lemma:parallelism} and Lemma~\ref{lemma:centralization} imply Theorem~\ref{thm:optimal_scheme}.

\subsection{Proof of Theorem 2}
\label{app:proof_th2}

\begin{customthm}{2}[Load Balance of Algorithm~\ref{alg:hash}]
Given a dense tensor $G$ with $|G|$ parameters.
Algorithm~\ref{alg:hash} provides a mapping $f:I \to [n]$ such that
\begin{enumerate}[nosep]
    \item With probability at least $1-1/n$, its imbalance ratio of push is at most $1 + \Theta(\sqrt{\frac{n \log n}{|G|d_G}}) $. 
    \item With probability at least $1-1/n$, the imbalance ratio of pull is at most 
    $1 + \Theta(\sqrt{\frac{n \log n}{|G|d_G^n}}) $.
\end{enumerate}
\end{customthm}

\begin{proof}
 The imbalance ratio of Algorithm~\ref{alg:hash} is only determined by $h_0:\mathbb{N}_{+}\to[n]$, while the hash function set $H$ focuses on exact memory write.

The number of indices in $I_i$ is $|G|d_G$, where $d_G$ is the density of $G$.
Since $h_0$ is data-independent, part 1 in Problem~\ref{prob:load_balance} can be formulated as: given $|G|d_G$ balls, we would like to toss them into $n$ bins with the universal hash function $h_0$. 
Taking the results from \cite{azar1994balanced}, the maximum load of the bins is at most $\frac{|G|d_G}{n}+\Theta(\sqrt{|G|d_G \log n/n})$ with probability at least $1-1/n$. 
Recall the definition of the imbalance ratio of push in Definition~\ref{def:imbalance_ratio}:
\begin{align*}
    \small
    Push_{h_0}^n = \max_{i, j \in [n]} \frac{n|I_i^j|}{|I_i|}.
\end{align*}
Because $\max \{|I_i^j|\} \le \frac{|G|d_G}{n}+\Theta(\sqrt{|G|d_G \log n/n})$, we have
\begin{align*}
    \small
    Push_{h_0}^n & \le  \frac{|G|d_G + \Theta(\sqrt{|G|d_G n\log n}) }{|G|d_G} \\
             & = 1 + \Theta(\sqrt{\frac{n \log n}{|G|d_G}}),
\end{align*}
with probability at least $1-o(1)$.
Thus, we finish the proof of the first part.

Since $h_0$ is data-independent, part 2 in Problem~\ref{prob:load_balance} can be formulated as: given $|I| = |G|d_G^n$ balls, we would like to toss them into $n$ bins with the universal hash function $h_0$.
The maximum load of the bins is at most $\frac{|G|d_G^n}{n}+\Theta(\sqrt{|G|d_G^n \log n/n})$ with probability at least $1-1/n$.
Recall the definition of the imbalance ratio of pull in Definition~\ref{def:imbalance_ratio}:
\begin{align*}
    \small
    Pull_{h_0}^n = \max_{i \in [n]} \frac{n|I_i^{'}|}{|I|}.
\end{align*}
Because $\max \{|I_i^{'}|\} \le \frac{|G|d_G^n}{n}+\Theta(\sqrt{|G|d_G^n \log n/n})$, we have
\begin{align*}
    \small
    Pull_{h_0}^n & \le  \frac{|G|d_G^n + \Theta(\sqrt{|G|d_G^n n\log n}) }{|G|d_G^n} \\
             & = 1 + \Theta(\sqrt{\frac{n \log n}{|G|d_G^n}}),
\end{align*}
with probability at least $1-1/n$.
Thus, we finish the proof of the second part.

\end{proof}

\subsection{Proof of The Property of Hash Bitmap}
\label{app:proof_th3}

In Pull of {\name}, the total hash bitmap size received at each worker from all servers is constantly $|G|/32$.

\begin{proof}
    Suppose there are $n$ servers.
    The set of indices that should be pushed to Server $i$ is $\mathbb{I}_i = \{idx \in [1, |G|] ~| ~h_0(idx) = i\}$.
    With \texttt{hash\_bitmap\_encode()}, the size of the hash bitmap encoded at Server $i$ is $|\mathbb{I}_i| / 32$.
    Because each worker needs to receive the hash bitmap from all the servers, the total size is $\sum_{i=0}^{n-1}|\mathbb{I}_i|/32 = |G| / 32 $.
\end{proof}
%%%%%%%%%%%%%%%%%%%%%%%%%%%%%%%%%%%%%%%%%%%%%%%%%%%%%%%%%%%%%%%%%%%%%%%%%%%%%%%%
\end{document}